%% file: main.tex
\documentclass[accepted]{uai2021} % after acceptance, for a revised
\usepackage[american]{babel}
\usepackage{natbib} % has a nice set of citation styles and commands
\usepackage{mathtools} % amsmath with fixes and additions
\usepackage{booktabs} % commands to create good-looking tables
\usepackage{tikz} % nice language for creating drawings and diagrams

\usepackage{import} % for importing sections

\usepackage{xcolor}

\usepackage{amsmath, amssymb}

\newcommand{\Pawandone}[1]{\textcolor{blue}{}}

\title{Generating Adversarial Examples with Graph Neural Networks}

\author{Florian Jaeckle and M. Pawan Kumar\\
Department of Engineering Science\\
University of Oxford\\
\texttt{\{florian,pawan\}@robots.ox.ac.uk}}

\usepackage{UAI_preamble}

\begin{document}
\maketitle

\import{Sections/}{Abstract.tex}

\import{Sections/}{Introduction.tex}

\import{Sections/}{Related_Work.tex}
\import{Sections/}{Background.tex}

\import{Sections/}{GNN_Framework.tex}

\import{Sections/}{GNN_Training.tex}

\import{Sections/}{Dataset.tex}

\import{Sections/}{Experiments.tex}

\import{Sections/}{Conclusion.tex}

% \begin{acknowledgements} % will be removed in pdf for initial submission,
%                          % so you can already fill it to test with the
%                          % ‘accepted’ class option
%     Briefly acknowledge people and organizations here.

%     \emph{All} acknowledgements go in this section.
% \end{acknowledgements}
\clearpage
% \bibliography{uai2021-template}
\bibliography{UAI_bib}
\clearpage
\appendix

\onecolumn
\import{Sections/}{Appendix.tex}

\end{document}

%% file: Sections/Abstract.tex
\begin{abstract}
% \begin{itemize}
    % \item TODO
    % call our method advGNN or GNN-attack?
    % \item define intermediate bounds properly
    % \item change every y' to ytar
    % \item double check with journal submission
% \end{itemize}

    Recent years have witnessed the deployment of adversarial attacks to evaluate the robustness of Neural Networks. Past work in this field has relied on traditional optimization algorithms that ignore the inherent structure of the problem and data, or generative methods that rely purely on learning and often fail to generate adversarial examples where they are hard to find.
    To alleviate these deficiencies, we propose a novel attack based on a graph neural network (GNN) that takes advantage of the strengths of both approaches; we call it AdvGNN.
    Our GNN architecture closely resembles the network we wish to attack. 
    During inference, we perform forward-backward passes through the GNN layers to guide an iterative procedure towards adversarial examples.
    During training, its parameters are estimated via a loss function that encourages the efficient computation of adversarial examples over a time horizon. 
    We show that our method beats state-of-the-art adversarial attacks, including PGD-attack, MI-FGSM, and Carlini and Wagner attack, reducing the time required to generate adversarial examples with small perturbation norms by over 65\%. Moreover, AdvGNN achieves good generalization performance on unseen networks.
    Finally, we provide a new challenging dataset specifically designed to allow for a more illustrative comparison of adversarial attacks.

\end{abstract}

%% file: Sections/Introduction.tex
\section{Introduction}

Ever since \cite{szegedy2013intriguing} showed that Neural Networks (NNs) are susceptible to adversarial attacks, it has become common practice to evaluate their robustness to various types of adversarial attacks. Most attack schemes use standard techniques from the optimization literature without significant adaptation for the specific problem at hand
\citep{szegedy2013intriguing, moosavi2017universal, goodfellow2014explaining, madry2018towards, papernot2016limitations}.
% \Pawandone{Citations needed.}
%They either aim to find an adversarial example that is misclassified with the highest level of confidence given an allowed perturbation or they try to find the minimum amount of perturbation needed to find an adversarial example.
% \Pawandone{Is the above sentence necessary in the Intro section?}
At the other end of the spectrum are purely machine learning based techniques, which aim to learn the underlying probability distribution of adversarial perturbations to generate adversarial examples 
\citep{baluja2017adversarial, zhao2017generating, poursaeed2018generative, song2018constructing}.
% \Pawandone{Citations needed.}
% However, if the adversarial probability distribution is small and adversarial examples are hard to find, non-iterative methods such as most generative attacks, fail to match the performance of optimization-based iterative methods.
However, the inductive bias incorporated in the network architectures of generative models ignores the iterative structure of optimization-based attacks. As a result, generative models often fail to match the performance of iterative optimization-based methods on finding minimal perturbations leading to adversarial examples.
% \Pawandone{Criticism needs to be more specific.}
We therefore introduce a novel attacking method that combines the optimization based approach with learning. 
% \Pawandone{Cannot name your method without introducing a GNN ... save for later.}

Specifically, we propose the use of a graph neural network (GNN) that assists an iterative procedure resembling standard optimization techniques. The architecture of the GNN closely mirrors that of the network we wish to attack.
% \Pawandone{`attack' not `verify'}
Given an image, its true class and an incorrect target class, at each iteration the GNN proposes a direction for potentially maximizing the difference between the logits of the incorrect class and the correct class. Henceforth, we refer to the objective function we wish to maximize via the GNN as the adversarial loss function.
Every single evaluation of the GNN is made up of one or more forward and backward passes that mimic a run of the network that we are attacking.
% \Pawandone{Again, not `verifying', but `attacking' ... change throughout the paper.}
When training the GNN we consider a horizon with a decay factor to output a direction of movement that maximizes the adversarial loss function.
% \Pawandone{Avoid using `ascent' .. there is no guarantee that the direction outputted but the GNN will actually be an ascent direction}
By using a parameterization of the GNN that depends only on the type of neurons and layers and not on the underlying architecture, we can train a GNN using one network and test it on another.

Our other main contribution is introducing a new method to assess the strength and efficiency of adversarial attacks. In the literature adversarial attacks are often compared using a trained model and some fixed allowed perturbation size.
% \Pawandone{Only makes sense to explicitly say `second' if you have earlier explicitly said `first'}
The method that manages to find an adversarial example for the highest number of images is considered to be the strongest one.
However, the network to be attacked is often robust for a significant proportion of images. All attacks on these images will therefore fail. Conversely, for other images adversarial perturbations are very easy to find, again not demonstrating significant differences between methods. We therefore introduce a challenging dataset on three different neural networks of different sizes that are solely made up of properties for which adversarial examples exist. The size of the allowed perturbation is deliberately chosen for each element in the dataset leading to a very high level of difficulty.
% \Pawandone{Can we say something here about the difficulty of finding adversarial examples in this new dataset.}
We hope that providing this new dataset will allow for a more efficient and meaningful comparison of different adversarial attacks in the future.
% \Pawandone{Would prefer to call the properties on all three networks as just a single dataset, instead of three different datasets.}

We compare our method, which we call AdvGNN, against various attacks on this dataset. AdvGNN reduces the average time required to find adversarial examples by more than 65\% compared to several state-of-the-art attacks and also significantly reduces the rate of unsuccessful attacks. AdvGNN also achieves good generalization performance on unseen larger models.

% TODO results

% TODO github link once finished

\iffalse
Skip 1-2\\
don't mention adv training\\
common practice to evaluate NNs with adv attacks\\
normally based on optimization - fails to exploit inherent structure of the problem and the data - there are purely generative (too extreme to rely purely on learning). take template of opt and assist with learning\\
skip 4\\
5 very brief

\begin{enumerate}
    \item Deep Learning and Neural Networks
    \item often susceptible to Adversarial Attacks
    \item what adv attacks are used for: adversarial training, evaluating robustness 
    \item types of current attacks (targeted vs untargeted) (white box vs black box) (universal vs image-dependent)
    \item optimization based (e.g. CW) vs sensitive analysis (e.g. pgd) vs generative (e.g. advGAN)
    \item aim to combine the learnt element of generative method with optimization
    \item high-level description of GNN
    \item describe our dataset - new way to compare adversarial attacks
    \item quick summary of results
    \item (perhaps include a visual example of an adversarial example returned by the GNN)
\end{enumerate}

\fi

%% file: Sections/Related_Work.tex
\section{Related Work}

In this work we focus on white-box image-dependent targeted attacks, the strongest form of adversarial attacks.

Studying adversarial attacks, and white-box attacks in particular, has become an active field of research over the last few years. Adversarial attacks can be separated into three main categories \citep{serban2020adversarial}. One class of attacks aims to find an adversarial example that lies within some allowed perturbation and  that the network misclassifies with a high level of confidence.
% \Pawandone{Why the `first' class?}
 \cite{goodfellow2014explaining} proposed the Fast Gradient Sign Method (FGSM) that takes a single step towards the gradient of the adversarial loss function.
The Iterative Fast Gradient Method (I-FGSM) \citep{kurakin2016adversarial} and Projected Gradient Attack (PGD) \citep{madry2018towards} both apply FGSM iteratively, taking several steps towards the sign of the gradient.
% \Pawandone{What does `steepest' gradient mean here? Do we mean `signed' gradient?}
\cite{dong2018boosting} proposed adding momentum to I-FGSM, thus significantly improving its performance (MI-FGSM).
% \Pawan{Needs a critique here.}

A similar line of research aims to find an adversarial example with the smallest possible perturbation. 
\cite{szegedy2013intriguing} proposed using limited-memory box constrained optimization (BFGS) to find the smallest perturbation required to change the prediction of the network. \cite{carlini2017towards} approximate the objective function using a simpler linear function that can be solved using standard optimization algorithms. \cite{moosavi2016deepfool} introduced the Deepfool attack that exploits the assumption that the network behaves linearly near the original input.
Both of these types of attacking strategies ignore the rich inherent structure of the problem and the data, information that can be used to come up with better ascent directions.
% \Pawandone{Needs a critique here.}

A third class of attacks includes generative methods. 
\cite{baluja2017adversarial} train a second neural network (ATN) that, given an input, aims to output an adversarial example.
\cite{poursaeed2018generative} trained a generative method that also learns to generate image-specific perturbations. 
\cite{xiao2018generating} propose the use of a GAN that learns to approximate the distribution of the original images.
All of these methods ignore the iterative nature of many optimization algorithms, resulting in a lower success rate in generating adversarial examples that are very close to original images.
% \Pawandone{Needs a critique here.}

We propose using a Graph Neural Network (GNN) to combine the strengths of both the optimization based and learning based methods to generate adversarial examples more efficiently.
% \Pawandone{Why `better'?}
GNNs have been used in Neural Network Verification to learn the branching strategy in a Branch-and-Bound algorithm \citep{Lu2020Neural} and to estimate better bounds \citep{dvijotham2018training, gowal2019dual}, but to the best of our knowledge they have not yet been used to generate adversarial examples.
We show in this work how they can be employed successfully for this task.

%% file: Sections/Background.tex
\section{Problem Definition} 
In this section we define the problem of finding adversarial examples, and outline some of the most popular approaches to solving it.

We are given a neural network $f:\mathbb{R}^d \mapsto \mathbb{R}^m$ that takes a $d$-dimensional input and outputs a confidence score for $m$ different classes.
\Pawandone{Use $\mathbb{R}$ etc. from the asmsymb package instead of $R$ etc. throughout the paper.}
\Pawandone{Don't like the use of lower case for one dimension and upper case for the other one.}
Specifically, we consider an $L$ layer feed-forward neural network, with non-linear activations $\sigma$ such that for any $\x_0\in {\cal C} \subseteq {\mathbb{R}}^{d}$, $f(\x_0)= \hat{\x}_{L}\in {\mathbb{R}^m}$, where
\Pawandone{$C$ is overloaded}
\label{eq:nn-formula}
\begin{align}
        \hat{\x}_{i+1} &= W^{i+1}\x_{i} + \bb^{i+1}, \qquad &&\text{for }i=0,\dots, L-1, \label{eq:linear} \\
        \x_{i} &= \sigma(\hat{\x}_i), \qquad &&\text{for }i=1,\dots, L-1. \label{eq:relu}
\end{align}
The terms $W^{i}$ and $\bb^{i}$ refer to the weights and biases of the $i$-th layer of the neural network f, and ${\cal C}$ is some convex input domain. Every convolutional filter can be rewritten as a linear layer; hence for the sake of clarity we treat convolutional layers like we do linear ones.
%In our case, we use the ReLU activation defined as $\sigma(x) = \max(x, 0)$ as it is widely used in machine learning in general \citep{krizhevsky2012imagenet, maas2013rectifier}.
%
Given an image $\x$, its true class $y$, an incorrect class $\ytar$, and an allowed perturbation $\epsilon$, a targeted attack aims to find $\x'$, such that
\begin{equation}\label{eq:adv_example}
    f(\x')_{\ytar} \ge f(\x')_{y} \;\text{ and } \; d(\x, \x') \leq \epsilon,
\end{equation}
for some distance measure $d$. In other words we aim to find an adversarial example $\x'$ that is close to the original input but is misclassified as $\ytar$. 
Problem \eqref{eq:adv_example} is often reformulated as follows:
\begin{equation}\label{eq:adv_loss}
    max_{\x' \in \normball} \; \advloss = f(\x')_{\ytar} - f(\x')_{y}, 
\end{equation}
where $\normball$ is an $\epsilon$-sized norm-ball around $\x$, that is, 
\begin{equation}
    \normball := \{x' \mid d(\x, \x') \leq \epsilon\}.
\end{equation}
We refer to $L$ as the adversarial loss from now on. If $\advloss \ge 0$ then $\x'$ is considered an adversarial example.

FGSM \citep{goodfellow2014explaining}, a fast attack on the $l_\infty$ norm, aims to solve $\eqref{eq:adv_loss}$ by using the sign of the gradient of the adversarial loss: 
\begin{equation}\label{FGSM}
    \x' = \x + \epsilon \sgn(\nabla_\x \advloss).
\end{equation}

\cite{madry2018towards} proposed applying this step iteratively, which equates to running Projected Gradient Descent (PGD) on the negative adversarial loss:
\begin{equation}\label{PGD}
    \x^{t+1} = \Pi_{\normball} \left( \x^t + \alpha \sgn(\nabla_\x \advloss \right).
\end{equation}

Using the sign of the gradient of the adversarial loss as the direction of movement is effective when we don't have access to more information about the problem. However, we argue that in the white-box setting, where we have access to more information, the effectiveness of this approach is limited.
We aim to replace the gradient by a more informed direction that, along with the gradient, takes the inherent structure of the problem and the data into consideration.

\iffalse
Many other methods to estimate \ref{eq:adv_loss} have been propose.

should become a section.
explain PGD attacks
aim to replace the direction of movement. typically given by gradient (makes sense in general optimization methods when we don't know much else)
replace with more informed direction
\subsection{Problem Definition}

Relatively straight forward. Define targeted adversarial $l_\infty$ attack and how to rewrite it as $f(x) \leq 0$

\fi

%% file: Sections/GNN_Framework.tex
% \section{GNN Framework}

% Similar to JMLR paper

% We need to describe the lp we are solving to then use as features as well as the Kolter Wong method. Perhaps this should be moved to the appendix, as it's not that relevant to the `main story` of the paper

\section{GNN Framework}\label{sec:GNN_framework}
The key observation of our work is that several previously known attacks can be thought of as performing forward-backward style passes through the network to compute an ascent direction for the adversarial loss function. Examples include, PGD, I-FGSM, and \CW (the method proposed by \cite{carlini2017towards}). However, the exact form of the passes is restricted to those suggested by standard optimization algorithms, which are agnostic to the special structure of adversarial attacks. This observation suggests a natural generalization: parameterize the forward and backward passes, and estimate the parameters using a training dataset so as to exploit the problem and data structure more successfully. In what follows, we first provide an overview of our approach that achieves this generalization through graph neural networks (GNN). The remaining subsections describe the various components of the GNN and the forward and backward passes in greater detail.
 
 \subsection{Overview}
 
%  Inspired by the approach of \citet{Lu2020Neural}, who use a graph neural network (GNN) to come up with better branching decisions, we propose to use a GNN for efficient computation of adversarial examples.
 We propose to use a GNN for the efficient computation of adversarial examples.
 % maybe cite bounding paper once published as it's more relevant to our work
 Since previous attacks perform forward and backward passes on the network they wish to attack, it makes sense to use a GNN that mimics the architecture of that network as closely as possible.
 To this end, we treat the neural network as a graph $G_{NN}=(V_{NN}, E_{NN})$ and provide it as input for the GNN. 
 We denote the GNN as an isomorphic graph to $G_{NN}$, that is, $G_{GNN} = (V_{GNN}, E_{GNN})$ where there is a one-to-one correspondence between the nodes $V_{NN}$ and $V_{GNN}$, and edges $E_{NN}$ and $E_{GNN}$.
 For every node $v \in V_{GNN}$ we first compute a feature vector $\feat$, which contains local information about the node. We then use this feature vector and a learned function $g$ to compute an embedding vector $\mmu$. The high-dimensional embedding vector encapsulates a lot of the important information about the corresponding node, the structure of the neural network, and the state of the optimization algorithm. 
 The embedding vectors are initialized based on the node features and then updated using forward and backward passes in the GNN.
 Exchanging information with its neighbours ensures that the embedding vectors capture the global information of the structure of the problem.
 Once we have gotten a learned representation of each node we will convert the embedding vectors into a direction of movement.
 \Pawandone{What `dual variables'?}
 Having provided an overview we will now describe the GNN's main elements in greater detail.
 
 \subsection{GNN Components}

 \paragraph{Nodes.}
 We create a node $\vv_k[i]$ in our GNN for every node in the original network, where $k$ indexes the layer and $i$ the neuron.
 \Pawandone{If $k$ indexes the layer and $i$ indexes the neuron, then state it explicitly here.}
 We denote the set of all nodes in the GNN by $V_{GNN}$.

%  \paragraph{Node Features.}
%  For each node $\vv_k[i]$ we define a corresponding $q$-dimensional feature vector $\fki \in \mathbb{R}^q$
%  describing the current state of that node. 
%  \Pawandone{$d$ is overloaded.}
%  We define the node features for the input layer as follows: 
%  \begin{equation}\label{eq:feature_input}
%     \fkO := \left(\x^{t}, \nabla_\x L(\x^t, y, y'), \lb_k[i], \ub_k[i], \z^{lp}_k[i]\right)^{\top},
%  \end{equation}
%  and for the hidden and final layers as:
%   \begin{equation}\label{eq:feature_inter}
%     \fki := \left(\lb_k[i], \ub_k[i], \rrho_k[i] \right)^{\top}.
%  \end{equation}
%  Here, $\x^{t}$ is our current point, $\nabla_\x L(\x, y, y')$ is the gradient at the current point, and $\lb_k[i]$, and  $\ub_k[i]$ are the bounds for each node.
%  \Pawandone{Can't start two consecutive sentences with `Here'}
%  Further, $\rrho_k$ is the current assignment to the corresponding dual variables computing using the method by \cite{wong2017provable} and $\z^{lp}_k[i]$ is the input corresponding to the primal solution of the dual (a more detailed explanation of the dual problem can be found in the Appendix \eqref{app:kolter_wong}).
%  \Pawandone{This is a bit confusing ... maybe it's best to avoid the details altogether and push them to the appendix. Something more high-level can be said here. Let's discuss during the meeting.}
%  While more complex features could be included, we deliberately chose the simple features described above and rely on the power of GNNs to efficiently compute an accurate ascent direction.

\paragraph{Node Features.}
For each node $\vv_k[i]$ we define a corresponding $q$-dimensional feature vector $\fki \in \mathbb{R}^q$ describing the current state of that node. Its exact definition depends on the task we want to solve. In our experiments the feature vectors consist of three parts:
the first part captures the gradient at the current point;
the second part includes the lower and upper bounds for each neuron in the original network based on the bounded input domain;
and the third part encapsulates information that we get from solving a standard relaxation of the adversarial loss from the incomplete verification literature.
A more detailed analysis can be found in Appendix \ref{app:GNN_architecture}.

%We first compute the intermediate bounds of each node in the network $(\lb, \ub)$ using the WK method~\citep{wong2017provable} which is explained in greater detail in Appendix \eqref{app:kolter_wong}. Intermediate bounds indicate all values that a hidden node can take given a bounded input domain.  As WK computes the intermediate bounds by solving a dual problem we also get a set of dual variables $({\rrho})$ and a primal solution $(\x_{lp})$ which we use as features.

While more complex features could be included, we deliberately chose the simple features described above and rely on the power of GNNs to efficiently compute an accurate direction of movement.

 \paragraph{Edges.}
 We denote the set of all the edges connecting the nodes in $V_{GNN}$ by $E_{GNN}$.
 The edges are equivalent to the weights in the neural network that we are trying to attack. We define $e^k_{ij}$ to be the edge connecting nodes $v_k[i]$ and $v_{k+1}[j]$ and assign it the value of $W^k_{ij}$.
 \Pawandone{What about biases? How are they treated?}

\paragraph{Embeddings.}
For every node $v_k[i]$ we compute a corresponding $p$-dimensional embedding vector $\mmu_k[i] \in \RR^p$ using a learned function $g$:
\begin{equation}
    \mmu_k[i] := g (\fki).
\end{equation}
In our case $g$ is a simple multilayer perceptron (MLP), which is made up of a set of linear layers $\Theta_i$ and non-linear ReLU activations.
We have the following set of trainable parameters:
% \begin{equation}
%     \Theta_0 \in \RR^{q \times p}, \quad
%     \Theta_1, \dots, \Theta_{T_1} \in \RR^{p \times p}, \quad
%     \bb_0, \dots, \bb_{T_1} \in \RR^{p}.
% \end{equation}
\begin{equation}
    \Theta_0 \in \RR^{q \times p}, \quad
    \Theta_1, \dots, \Theta_{T_1} \in \RR^{p \times p}.
\end{equation}
% \begin{equation}
%     \thetainp_0 \in \RR^{q \times p}, \quad
%     \thetainp_1, \dots, \thetainp_{T_1} \in \RR^{p \times p}, \quad
%     \bb_0, \dots, \bb_{T_1} \in \RR^{p}.
% \end{equation}
% \begin{align}
% \begin{split}
%     &\thetainp_0 \in \RR^{q' \times p}, \quad
%     \Theta_0 \in \RR^{q \times p} \; \\
%     &\thetainp_1, \dots, \thetainp_{T_1}, \; \Theta_1, \dots, \Theta_{T_1}  \in \RR^{p \times p}
% \end{split}
% \end{align}

Given a feature vectors $\feat_k$, we compute the following set of vectors:
% \begin{align}
%     \mmu^0_0 = \relu(\thetainp_0 \cdot \feat_0), \quad
%     \mmu^{l+1}_0 = \relu(\thetainp_{l+1} \cdot \mmu^l_0),\\
%     \mmu^0_k = \relu(\Theta_0 \cdot \feat_k), \quad
%     \mmu^{l+1}_k = \relu(\Theta_{l+1} \cdot \mmu^l_k), \\
%     {\quad \text{ for } k = 1, \dots, L-1}, \; { l = 1, \dots, T_1-1}.\nonumber
% \end{align}
% \begin{equation}
%     \mmu^0_k = \relu(\Theta_0 \cdot \feat_k + \bb_0), \quad
%     \mmu^{l+1}_k = \relu(\Theta_{l+1} \cdot \mmu^l_k + \bb_l).
% \end{equation}
\begin{equation}
    \mmu^0_k = \relu(\Theta_0 \cdot \feat_k), \quad
    \mmu^{l+1}_k = \relu(\Theta_{l+1} \cdot \mmu^l_k).
\end{equation}
We initialize the embedding vectors to be $\mmu_k = \mmu_k^{T_1}$, where $T_1+1$ is the depth of the MLP.

\subsection{Forward and Backward Passes}
So far, the embedding vector $\mmu$ solely depends on the current state of that node and does not take the underlying structure of the problem or the neighbouring nodes into consideration. 
We therefore introduce a method that updates the embedding vectors by simulating the forward and backward passes in the original network.
The forward pass consists of a weighted sum of three parts: the first term is the current embedding vector, the second is the embedding vector of the previous layer passed through the corresponding linear or convolutional filters, and the third is the average of all neighbouring embedding vectors:
\begin{dmath}\label{eq:forward_pass}
    \mmu'_k[i] = \relu \left( \thetafor_1 \mmu_k[i] + \thetafor_2 \left(W_k \mmu_{k-1} + \bb_{k-1}\right)[i] + \thetafor_3 \left(\sum_{j \in N(i)} \mmu_{k-1}[j] / Q_{k+1}[j]\right)[i] \right).
\end{dmath}
% \Pawandone{What does `existing deep learning functions' mean? Do you mean `existing deep learning libraries'?}
Similarly, we perform a backward pass as follows:
\begin{dmath}\label{eq:backward_pass}
    \mmu_k[i] = \relu \left( \thetaback_1 \mmu'_k[i] + \thetaback_2 (W_{k+1}^{T} \left(\mmu'_{k+1} - \bb_{k+1}\right))[i] + \thetaback_3 \left(\sum_{j \in N'(i)} \mmu'_{k+1}[j] / Q'_{k+1}[j]\right)[i] \right).
\end{dmath}
Here $\thetafor_1, \thetafor_2, \thetafor_3, \thetaback_1, \thetaback_2, \thetaback_3 \in \RR^{p \times p}$ are all learnable parameters and $W$ and $b$ are the weights and biases of the target network as defined in equations \eqref{eq:linear} and \eqref{eq:relu}. 
% The second term in equation \eqref{eq:forward_pass} and \eqref{eq:backward_pass} corresponds to passing the embedding vectors of the previous layer through the linear layer, whereas the third term takes the average of the embedding vectors of the previous layer.
Both \eqref{eq:forward_pass} and \eqref{eq:backward_pass} can be implemented using existing deep learning libraries.
To ensure better generalization performance to unseen neural networks with a different network architecture we include normalization parameters $Q$ and $Q'$. These are matrices whose elements are the number of neighbouring nodes in the previous and following layer respectively for each node.
We repeat this process of running a forward and backward pass $T_2$ times. The high-dimensional embedding vectors are now capable of expressing the state of the corresponding node taking the entire problem structure into consideration as they are directly influenced by every other node, even if we set $T_2 = 1$.

\subsection{Update Step}
Finally, we need to transform the $p$-dimensional embedding vector of the input layer to get a new direction $\dir$. We simply use a linear output function $\thetascore$ to get: 
\begin{equation}
    \dir = \thetascore \cdot \mmu_0.
\end{equation}
 Ideally the GNN would output a new ascent direction that will lead us directly to the global optimum of equation (\ref{eq:adv_loss}). However, as the problem is complex this may not be feasible in practice without making the GNN very large, thereby resulting in computationally prohibitive inference. Instead, we propose to run the GNN a small number of times to return directions that gradually move towards the optimum.
%  Given a step size $\eta^{t+1}$, our previous point $\x^t$, and the new ascent direction $\dir$ we update as follows: 
% \begin{equation}\label{GNN_update}
%     \x^{t+1} = \Pi_{x+S} \left( \x^t + \eta^t \sgn(\dir) \right).
% \end{equation}
%  Similar to many iterative optimization methods we decay our stepsize as we want to take smaller steps the closer we get to the optimal solution. Given an initial step size $\eta_0$, we define the step size at time $t$ as follows:  $\eta^t = \eta_{init} + (t/T) * (\eta_{fin} - \eta_{init})$

 Given a step size $\alpha$, our previous point $\x^t$, and the new  direction $\dir$ we update as follows: 
\begin{equation}\label{GNN_update}
    \x^{t+1} = \Pi_{\normball} \left( \x^t + \alpha \dir \right).
\end{equation}
 The hyper-parameters for the GNN computation of new directions of movement are the depth of the MLP ($T_1$), how many forward and backward passes we run ($T_2$), the embedding size ($p$), and the stepsize parameter $\alpha$. 
 \Pawandone{`duals'??}
 % parameters $\eta_{fin}$ and $\eta_{init}$.

%% file: Sections/GNN_Training.tex
\section{GNN Training}\label{sec:GNN_training}

% Note: l, u, p, x are not part of the dataset.
% Only x, y, y', eps, W are.
% 1. Run LP to get l, u, p, x
% 2. get random start point, sampled unified at random

Having described the structure of the GNN we will now show how to train its learnable parameters. 
Our training dataset ${\cal D}$ consists of a set of samples $d_i = \trainingsample$, each with the following components: a natural input to the neural network we wish to attack ($\x$), for example an image; the true class (y); a target class ($\ytar$);
the size of the allowed perturbation ($\epsilon$), which in our case is an $\ell_\infty$ ball; and the weights and biases of the neural network $(W, \bb)$. We note that the allowed perturbation can be unique for each datapoint.

In order to get the individual components that make up the feature vectors, we first compute the intermediate bounds of each node in the network using the method by \cite{wong2018provable} which is explained in greater detail in Appendix \ref{app:sec:inter_bounds}. We further solve a standard relaxation of the robustness problem via methods from the verification literature \eqref{app:sec:dual}.
% computes the intermediate bounds by solving a dual problem we also get a set of dual variables $({\rrho})$ and a primal solution $(\x_{lp})$ which we use as features.
% We first compute the intermediate bounds of each node in the network $(\lb, \ub)$ using the WK method~\citep{wong2017provable} which is explained in greater detail in Appendix \eqref{app:kolter_wong}. Intermediate bounds indicate all values that a hidden node can take given a bounded input domain.  As WK computes the intermediate bounds by solving a dual problem we also get a set of dual variables $({\rrho})$ and a primal solution $(\x_{lp})$ which we use as features.
Finally, we generate $s$ different starting points which we sample uniformly at random from the input domain $\normball$.
 
Recall that we do not use the GNN to directly compute the optimum adversarial example. Instead, we run it iteratively, where each iteration computes a new direction of movement.
In order for the training procedure to closely resemble its behaviour at inference time, it is crucial to train the GNN using a loss function that takes into account the adversarial loss across a large number of iterations $K$.

 Given the $i$-th training sample $d_i = \trainingsample \in {\cal D}$, and the $j$-th initial starting point we define the loss $\loss_{i, j}$ to be:
 \begin{equation}\label{loss_function}
     \loss_{i, j} = -\sum_{t=1}^K  L(\x^{i, j, t}, y^i, \ytar^i) * \gamma^t.
 \end{equation} 
 Instead of maximizing over the adversarial loss, we minimize over the negative loss.
 If the decay factor $\gamma \in (0,1)$ is low then we encourage the model to make as much progress in the first few steps as possible, whereas if $\gamma$ is closer to $1$, then more emphasis is placed on the final output of the GNN, sacrificing progress in the early stages.
 Readers familiar with reinforcement learning may be reminded of the discount rates used in algorithms such as Q-learning and policy-gradient methods.
 We sum over the individual loss values corresponding to each data point and each initial starting point to get the final training objective $\loss$:
\begin{equation}
        \loss = 
    \sum_{i=1}^{\mid D \mid} \sum_{j=1}^{s} \loss_{i, j}.
\end{equation}
In our experiments we train the GNN using the Adam optimizer \citep{kingma2014adam} and with a small weight decay.
\paragraph{Running Standard Algorithms using AdvGNN.}
As mentioned earlier, the motivation behind our GNN framework is to offer a parameterized generalization of previous attacks. We now formalize the generalization using the following proposition.
\begin{proposition}
    \label{prop:simulate_methods}
    AdvGNN can simulate FGSM \citep{goodfellow2014explaining}, PGD attack \citep{madry2018towards}, and I-FGSM \citep{kurakin2016adversarial}
    (proof in Appendix \ref{app:simulate_methods}).
    %(proof in appendix \ref{sec:app:proof_prop}).
\end{proposition}

%% file: Sections/Dataset.tex
\section{A New Dataset for comparing Adversarial Attacks}
In this section we describe our new dataset that has been specifically designed to compare state-of-the-art adversarial attacks.

% Say inspired by comparing defense methods/complete verification methods.

Previously, adversarial attacks were compared on how well they attack a trained neural network on a set number of images for a fixed allowed perturbation \citep{madry2018towards, dong2018boosting, carlini2017towards, moosavi2016deepfool}.
However, for many of the images there either does not exist an adversarial example in the allowed perturbed input space or there exist a large number of different adversarial examples. In the first case, we don't learn anything about the differences between different methods as none of them return an adversarial example, and for the latter case all attacks will terminate very quickly, again not providing any insights.
In practice only a small proportion of test cases affect the differences in performance between the various methods.

To alleviate this problem we provide a dataset where the allowed input perturbation is uniquely determined for every image in the dataset. This 
ensures that for every property there exist adversarial examples, but so few that only efficient attacks manage to find them.

We generate a dataset based on the CIFAR-10 dataset \citep{krizhevsky2009learning} for three different neural networks of various sizes. One which we call the \Base model, one with the same layer structure but more hidden nodes which we call the \Wide model, and one with more hidden layers which we refer to as the \Deep model.
All three are trained robustly using the methods of \cite{madry2018towards} against $l_\infty$ perturbations of size up to $\epsilon = 8/255$ (the amount typically considered in empirical works). Our dataset is inspired by the work of \cite{Lu2020Neural} who created a verification dataset to compare defense methods on the same three models. The different network architectures are explained in greater detail in Appendix \ref{app:network_architecutres}.

We generate the dataset by repeatedly picking an image from the CIFAR-10 test set, asserting that the network classifies the image correctly, and picking an incorrect class at random. We then aim to compute the smallest perturbation for which there exists an adversarial example by running an expensive binary search using PGD attacks with a large number of steps and restarts. A more detailed description of the algorithm can be found in Appendix \ref{app:generating_dataset}. We also generate a second dataset on the \Base model which we call the validation dataset and use to optimize various hyper-parameters for the attacks used in the next section.

Finally, we note that in the literature only the success rate is reported when comparing different methods. The time taken by different methods is not analysed and the efficiency of the attacks is thus sometimes hard to determine. We propose to compare methods by reporting the success rate over running time to show both the speed and the strength of adversarial attacks.

%% file: Sections/Experiments.tex
\section{Experiments}
We now describe an empirical evaluation of our method by comparing it to several state-of-the-art attacks on the CIFAR-10 dataset. We first outline the experimental setting (\S\ref{sec:setup}), before describing the attacks we compare our method to (\S\ref{sec:methods}), and finally analysing the results (\S\ref{sec:results}).

\subsection{ Setup }\label{sec:setup}
% \paragraph{Setup.}
\import{Figures}{cactus_plots.tex}

We run experiments on the dataset described in the previous section. The dataset is based on the CIFAR-10 dataset and includes three different networks to attack. All properties are SAT, meaning that there exists at least one adversarial example in the given input domain for each image and an overall success rate of 100\% is theoretically achievable. 
We use a timeout of 100 seconds for each property.
As most of the attacks we use rely on random initialisations the performance varies depending on the random seed. We thus run every experiment three times with three different seeds and report the average over the different runs.

All the experiments were run under Ubuntu 16.04.4 LTS. All attacks were run on a single Nvidia Titan V GPU and three i9-7900X CPUs each. The implementation of our model as well as all baselines is based on Pytorch \citep{paszke2017automatic}.

\subsection{Methods}\label{sec:methods}
% \paragraph{Methods.}
% \import{Tables}{main_exp_combined.tex}

We evaluate our methods by comparing it against PGD-Attack, MI-FGSM+, a modified version of MI-FGSM, and Carlini and Wagner attack, which according to several surveys on adversarial examples are all state-of-the-art methods \citep{akhtar2018threat, chakraborty2018adversarial, serban2020adversarial, huang2020survey}.

\paragraph{PGD.} The first baseline we run is PGD-attack \citep{madry2018towards}. As described before, PGD-attack picks an initial starting point uniformly at random and then iteratively performs Projected Gradient Descent on the negative adversarial loss \eqref{PGD}.
Based on an extensive hyper-parameter analysis (see Appendix $\ref{app:hparam_pgd}$) we pick the stepsize parameter $\alpha=0.1$, and set the number of iterations to $T=100$ . We perform random restarts until we have either managed to find an adversarial example or the time limit has been reached.

\paragraph{MI-FGSM+.} MI-FGSM is I-FGSM with an added momentum term.
MI-FGSM starts at the image $\x$ and takes $T$ steps of size $\epsilon/T$. Defining the stepsize as such ensures that the current point lies in the feasible region throughout the entire algorithm without the need to project. To strengthen the attack we perform random restarts as we do for PGD. To ensure that not all runs of MI-FGSM on the same image are identical we therefore have to choose the initial point randomly as well. Furthermore, we perform a hyper-parameter search not only over the momentum term $\mu$ and the number of iterations $T$ as done in the original paper, but also over the stepsize $\alpha$ (see Appendix \ref{app:hparam_mi-fgsm} for details). We run it with the following optimized parameters: $\alpha = 0.1$, $\mu = 0.5$, and $T=100$. This modified version of MI-FGSM is denoted as MI-FGSM$+$.

\paragraph{C\&W.} The third baseline we use is C\&W, the optimization-based attack proposed by \cite{carlini2017towards}. C\&W aims to find the smallest perturbation required to find an adversarial example by minimizing a loss function of the form $l(v) = c \cdot F(x + v) + \lVert (v - \tau)_+ \rVert_1$ for some surrogate function $F$, and constants $c$ and $\tau$. There are a total of six hyper-parameters which we optimize over on a validation dataset and which we describe in greater detail in Appendix $\ref{app:hparam_CW}$. We note that in the original implementation C\&W is often run until the minimum perturbation for which there exist at least one adversarial example is found. However, to be able to compare it to the other methods we stop the C\&W attack as soon as an adversarial example is found for the given perturbation value or when the time limit is reached.

\paragraph{AdvGNN.}
The final attack we run is AdvGNN. We train our AdvGNN on the \Base model and on 2500 images of the CIFAR-10 test set that are not part of the dataset we test on. The $\epsilon$ values which define the allowed perturbation for each training sample are computed in a similar procedure to the test datasets described above.
We train the GNN using the loss function described in section \S\ref{sec:GNN_training} with a horizon of 40 and with decay factor $\gamma = 0.9$.
The training loss function is minimized using the Adam optimizer \citep{kingma2014adam} with a weight decay of 0.001. The initial learning for Adam is 0.01, and is manually decayed by a factor of 0.1 at epochs 20, 30, and 35.
We pick the following values for the hyper-parameters of our AdvGNN: the stepsize $\alpha$ is 1e-2, the embedding size is $p=32$, and we perform a single forward and backward pass ($T_1 = T_2 = 1$).
To improve the performance on the \Deep model we fine-tune our AdvGNN for 15 minutes on the \Deep model before running the attack. Fine-tuning is run on 300 images that are not included in the \Deep test set. We use a fixed $\epsilon$ value of 0.25 for all images.

% As neither advGAN nor GAP beat Carlini-Wagner and advGNN outperforms Carlini-Wagner by an order of magnitude, we conclude that our methods also outperforms advGAN and GAP.
% Another advantage of our method compared to GAP for targeted attacks is that \cite{poursaeed2018generative} train a separate GAP for every class, whereas advGNN does not.

\subsection{Results}\label{sec:results}
% \paragraph{Results.}
\import{Figures}{cactus_plots_easy.tex}

\import{Tables}{base2.tex}
\paragraph{\Base Model.}
We run all four methods described in the previous section on the \Base model with a timeout of 100 seconds and record the percentage of properties successfully attacked as a function of time (Figure \ref{fig:base} and Table \ref{tab:base2}). \CW only manages to find an adversarial advantage for less than 5\% of all images. PGD outperforms \CW but still only manages to solve 17\% of all properties. MI-FGSM+ outperforms PGD, timing out on 26\% of all images with an average time of 40 seconds. AdvGNN beats all three methods reducing both the average time taken and the proportion of properties timed out on by more than 65\%.

\import{Tables}{wide2.tex}
\paragraph{\Wide Model.}
Next we compare the methods on the \Wide model  (Figure \ref{fig:wide} and Table \ref{tab:wide2}). AdvGNN has not seen this network during training and before running these experiments. MI-FGSM+ is again the best performing baseline, and AdvGNN the best performing method overall both in terms of average solving time and percentage of properties successfully attacked. AdvGNN reduces the time required to find an adversarial example by over 70\% compared to PGD and \CW, and by 20\% compared to MI-FGSM+. This demonstrates that AdvGNN achieves good generalization performance and can be trained on one model and used to run attacks on another.

\import{Tables}{deep2.tex}
\paragraph{\Deep Model.}
We also run experiments on the \Deep model (Table \ref{tab:deep2}, Figure \ref{fig:deep}). We remind the reader that the AdvGNN parameters have been fine-tuned on this model for 15 minutes to achieve better results. AdvGNN outperforms all three other attacks on this larger \Deep model both with respect to the total number of successful attacks and the average time of each attack. Figure \ref{fig:deep} shows that AdvGNN is still the best performing method even if we pick a shorter timeout of less than 100 seconds.

\paragraph{Easy Dataset.}
As some of the baselines, \CW in particular, struggle to successfully attack most of the properties in the previous experiment, we further compare the methods on a simpler dataset. We add a constant delta (0.001) to each epsilon value in the above dataset and reduce the timeout to 20 seconds. Increasing the allowed perturbation simplifies the task of finding an adversarial example as can be seen in Figure \ref{fig:cactus_easy_combined}. All methods manage to find adversarial examples more quickly than on the original dataset and time out on significantly fewer properties. The relative order of the methods is the same on all three models in both the original and the simpler dataset. %In particular AdvGNN outperforms the baselines on the \Base, the \Wide, and the \Deep models, reducing average solving time on the \Base model by at least 75\% compared to all baselines.
In particular, AdvGNN outperforms the baselines on all three models, reducing the percentage of unsuccessful attacks by at least 98\% on the \Base model and by more than 65\% on the \Wide and \Deep model.
We provide a more in-depth analysis of the results on the original and the easier dataset in Appendix \ref{app:further_results}.

% differences in performance are amplified and become clear
% \import{Tables}{Base_Wide_Deep.tex}
% \import{Tables}{easy_exp_combined.tex}

% \import{Tables}{easy_exp_seeds.tex}
% \import{Tables}{debug.tex}
% \import{Tables}{easy_exp_seeds.tex}

% We compare our method with previous work by showing how many properties are verified for any given amount of time in seconds 
% mention that MI-FGSM is 

% \newpage

% Every section (only one if we don't use ETH) should have 3 subsections. 1. dataset/setup (base model + never seen wide and deep model) 2. baselines/our method (including fine-tuning) 3. results

% if manage to do public datasets do them in a new group of 3 subsections

% \begin{enumerate}
%     \item Dataset Generation - how it's generated, what it looks like and why it's useful (make separate section)
%     \item explain baselines (PGD/I-FGSM, Mi-FGSM/PGD with momentum, Deepfool, Carlini-Wagner)
%     \item base model experiment
%     \item Wide model experiment
%     \item explain Fine-tuning and Deep model experiment
%     \item if i have enough time, include some public datasets
% \end{enumerate}

%% file: Figures/cactus_plots.tex
 \begin{figure*}[ht!]
	\centering
	\begin{subfigure}{.32\textwidth}
		\centering
		\includegraphics[width=\textwidth]{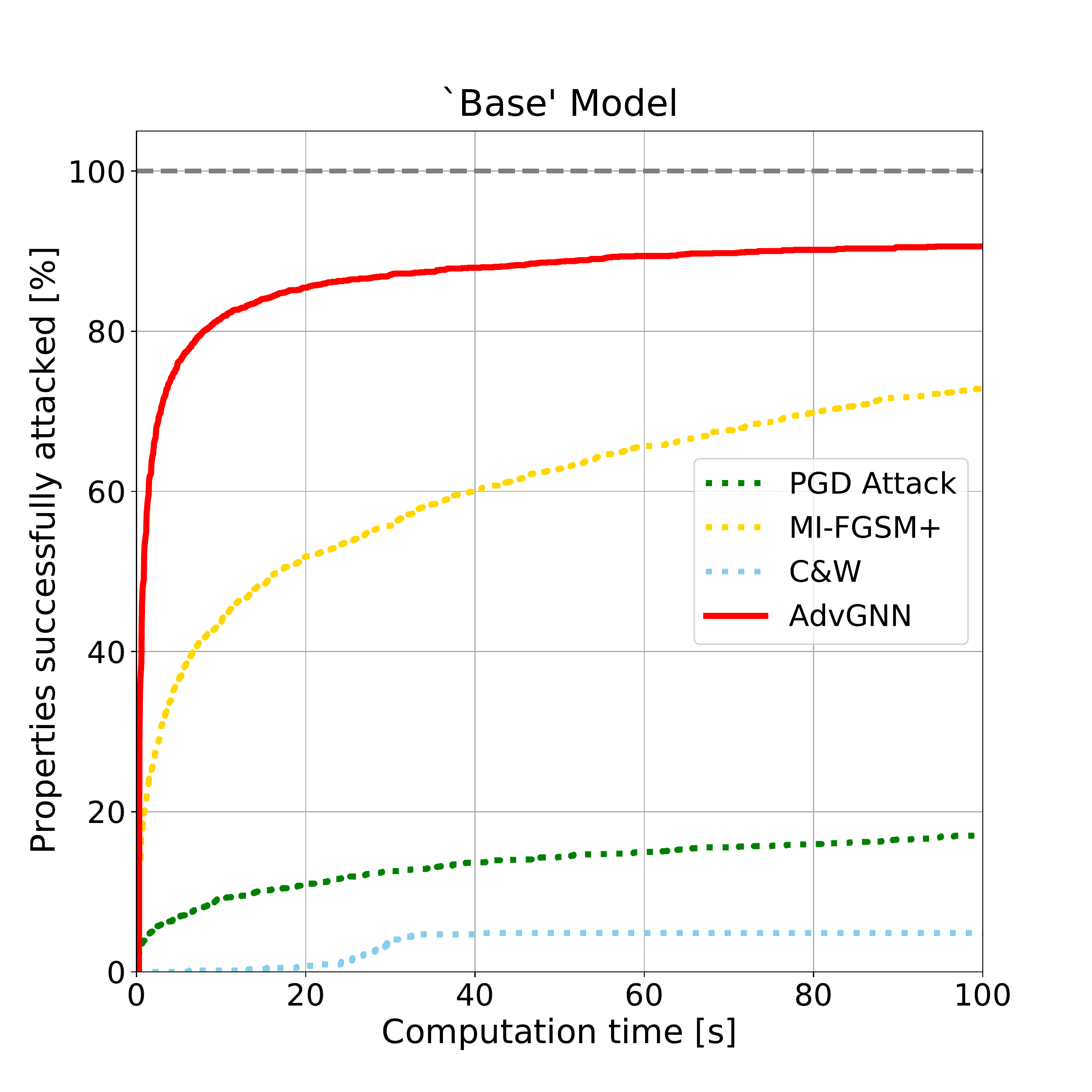}
		 \caption{\Base model}\label{fig:base}
	\end{subfigure}
	\begin{subfigure}{.32\textwidth}
		\centering
		\includegraphics[width=\textwidth]{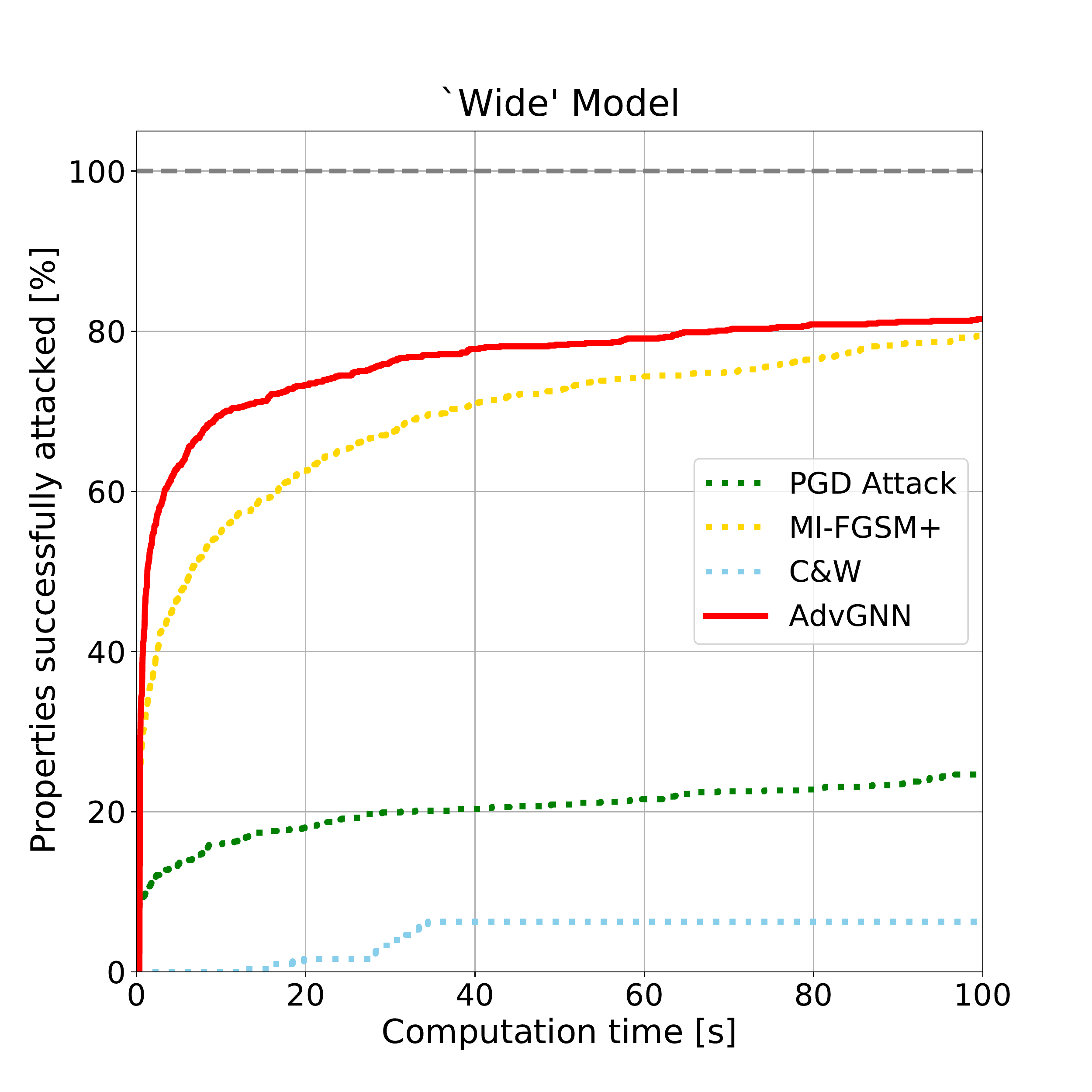}
		\caption{\Wide model}\label{fig:wide}
	\end{subfigure}
	\begin{subfigure}{.32\textwidth}
		\centering
		\includegraphics[width=\textwidth]{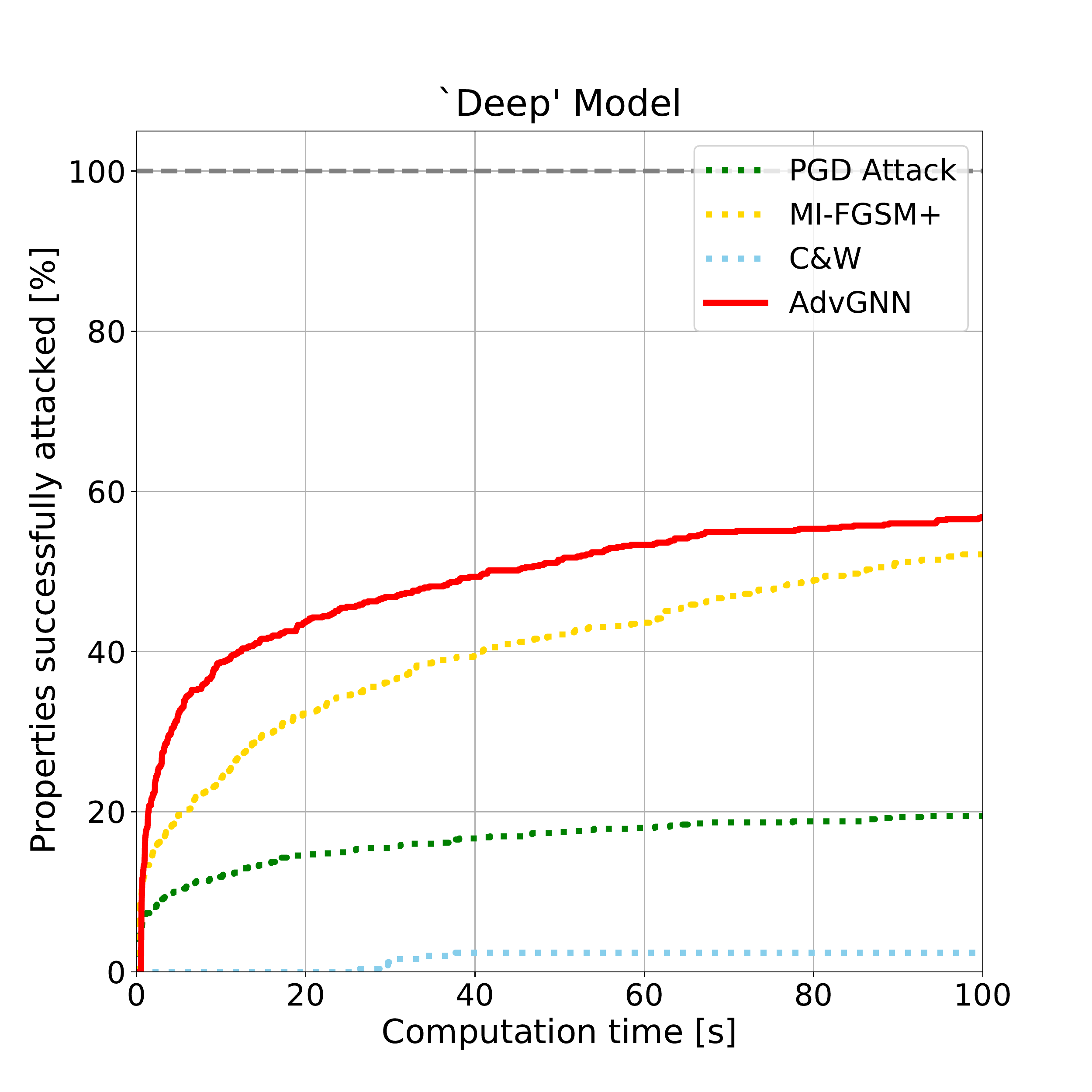}
		\caption{\Deep model}\label{fig:deep} 
	\end{subfigure}
	\caption{Cactus plots for experiments on the \Base model (left), \Wide model (middle) and the \Deep model (right). For each, we compare the different attacks by plotting the percentage of successfully attacked images as a function of runtime. Baselines are represented by dotted lines. AdvGNN beats all baselines on all three models for any chosen timeout value.}
	\label{fig:cactus_main_combined}
\end{figure*}

%% file: Figures/cactus_plots_easy.tex
 \begin{figure*}[ht!]
	\centering
	\begin{subfigure}{.32\textwidth}\label{fig:base_easy} 
		\centering
		\includegraphics[width=\textwidth]{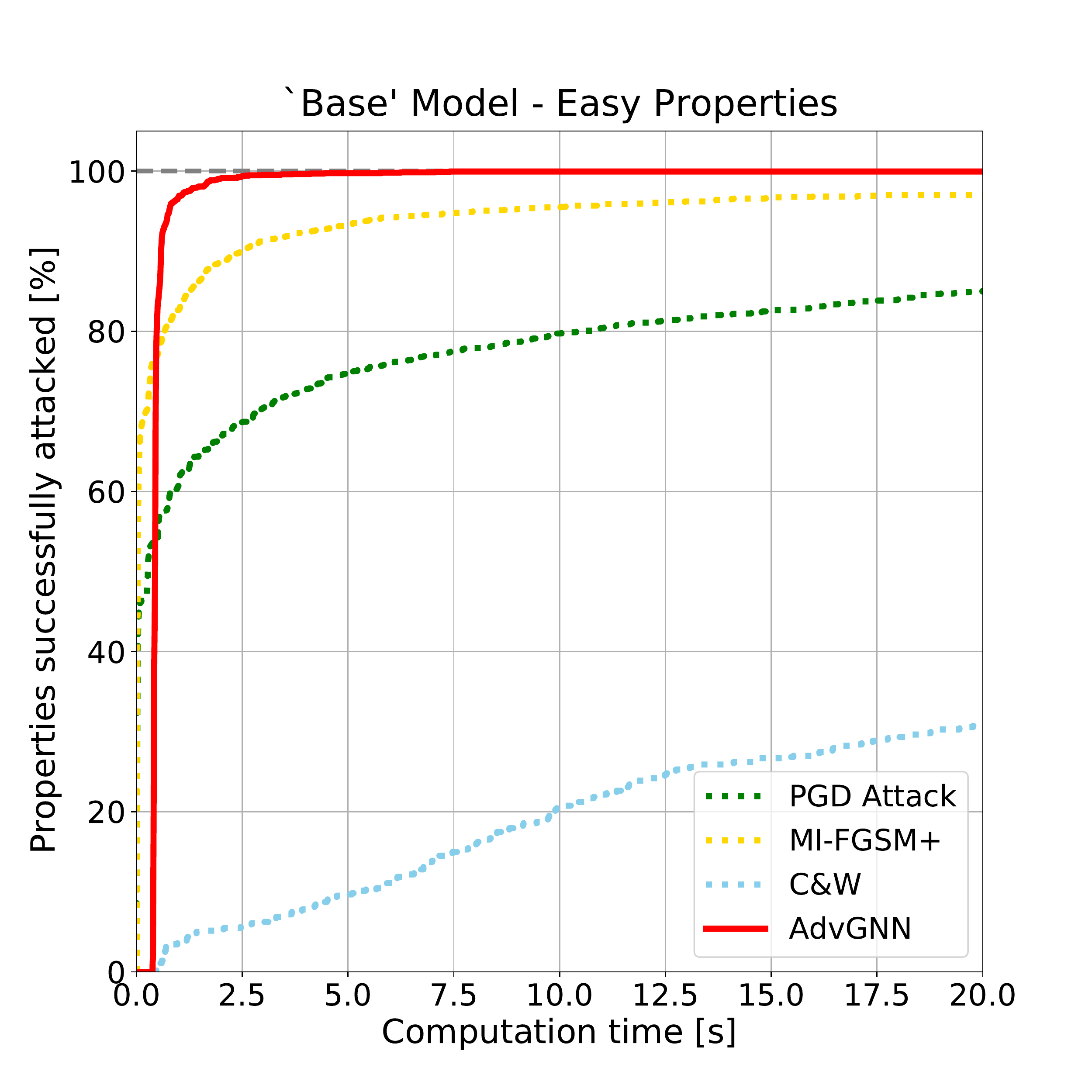}
		%\subcaption{subfigure}{680 steps of subgradient vs. 1040 steps of Dvijotham}
		%\label{fig:600prox}
	\end{subfigure}
	\begin{subfigure}{.32\textwidth}\label{fig:wide_easy} 
		\centering
		\includegraphics[width=\textwidth]{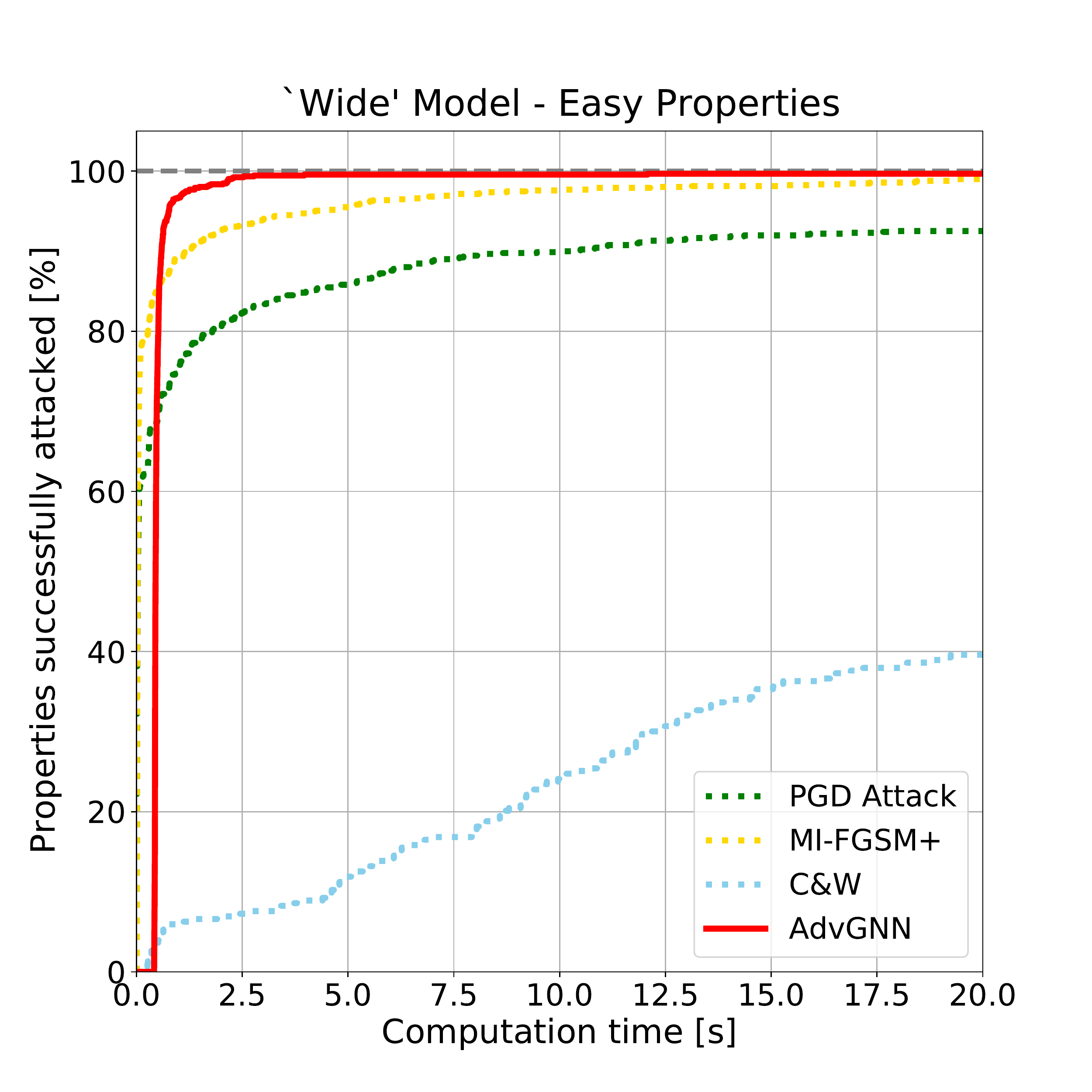}
		%\subcaption{400 steps of proximal vs. 680 steps of subgradient}
		%$label{fig:300prox} 
	\end{subfigure}
	\begin{subfigure}{.32\textwidth}\label{fig:deep_easy} 
		\centering
		\includegraphics[width=\textwidth]{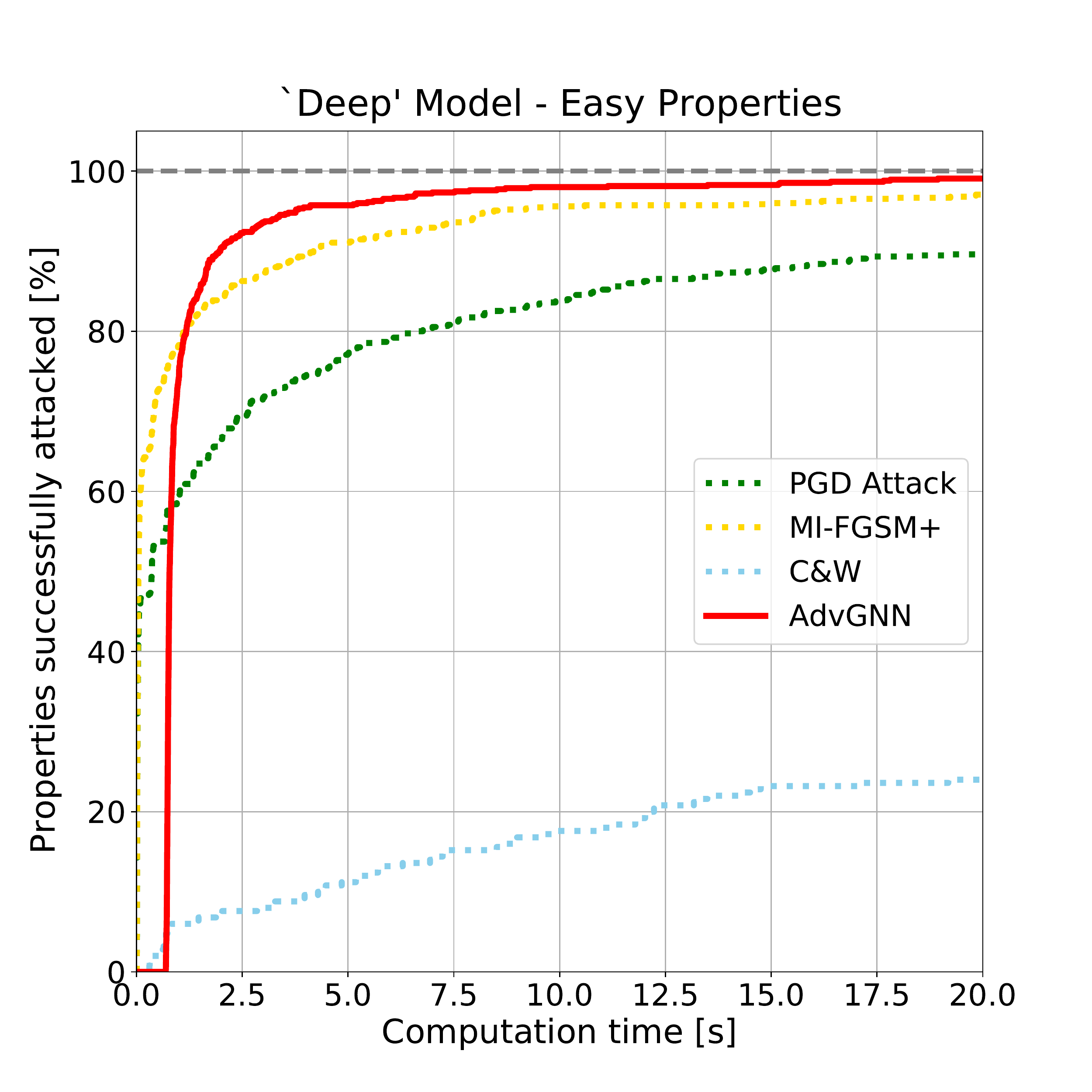}
		%\subcaption{400 steps of proximal vs. 680 steps of subgradient}
		%$label{fig:300prox} 
	\end{subfigure}
	\caption{Cactus plots for experiments on the easier version of the dataset on the \Base model (left), \Wide model (middle) and the \Deep model (right). We add a small constant $\delta=0.001$ to each perturbation size $\epsilon^i$.
    For each model, we compare the attacks by plotting the percentage of successfully attacked images as a function of runtime. AdvGNN is the best performing attack on all three models.}
	\label{fig:cactus_easy_combined}
\end{figure*}

%% file: Tables/base2.tex
\begin{table}[h]
    \centering
    \caption{\Base Model. We compare average (mean) solving time and the percentage of properties that the methods time out on when using a cut-off time of 100s.}\label{tab:base2}
    \begin{tabular}{rll}
      \toprule % from booktabs package
      \bfseries Method & \bfseries Time(s) & \bfseries Timeout(\%) \\
      \midrule % from booktabs package
 PGD Attack &   87.412 &      82.995        \\ 
 MI-FGSM+ &   40.438 &      27.145     \\
 C\&W &   97.385 &      95.164 \\
 AdvGNN &   \bfseries 13.527 &       \bfseries 9.412   \\
      \bottomrule % from booktabs package
    \end{tabular}
\end{table}

%% file: Tables/wide2.tex
\begin{table}[h]
    \centering
    \caption{\Wide Model. We compare the methods on the \Wide model.}\label{tab:wide2}
    \begin{tabular}{rll}
      \toprule % from booktabs package
      \bfseries Method & \bfseries Time(s) & \bfseries Timeout(\%) \\
      \midrule % from booktabs package
 PGD Attack &   80.415 &      75.358    \\ 
 MI-FGSM+ &   31.144  &      20.462     \\
 C\&W &   96.366 &      93.729 \\
 AdvGNN &   \bfseries 24.089 &       \bfseries 18.482   \\
      \bottomrule % from booktabs package
    \end{tabular}
\end{table}

%% file: Tables/deep2.tex
\begin{table}[h]
    \centering
    \caption{\Deep Model. We compare the different methods on the \Deep model.}\label{tab:deep2}
    \begin{tabular}{rll}
      \toprule % from booktabs package
      \bfseries Method & \bfseries Time(s) & \bfseries Timeout(\%) \\
      \midrule % from booktabs package
 PGD Attack &   84.349 &      80.533        \\ 
 MI-FGSM+ &    60.578 &       47.867    \\
 C\&W &  99.321 &      97.600 \\
 AdvGNN & \bfseries 51.669 & \bfseries 43.200   \\
      \bottomrule % from booktabs package
    \end{tabular}
\end{table}

%% file: Sections/Conclusion.tex
\section{Discussion}

We introduced AdvGNN, a novel method to generate adversarial examples more efficiently that combines elements from both optimization based attacks and generative methods. We show that AdvGNN beats various state-of-the-art baselines reducing the average time taken to find adversarial examples by between 65 and 85 percent. We further show that AdvGNN generalizes well to unseen methods.
Moreover, we introduced a novel challenging datasets for comparing different adversarial attacking methods. We show how it enables an illustrative comparison of different attacks and hope it will encourage the development of better attacks in the future.

Future work might include using AdvGNN for adversarial training, or for adversarial image detection. Furthermore, one could try incorporating AdvGNN into a complete verification method.

\iffalse
add quick GNN (trained on very few images)
train GNN without lp 
\fi

%% file: Sections/Appendix.tex
% \section{Network Architectures}\label{app:network_architecutres}
% \import{Tables}{network_architectures.tex}

% \section{Generating the Dataset}\label{app:generating_dataset}

% We generate a dataset for three different models: the \Base model, the \Wide model, and the \Deep model. For each of the three models we run the $GENERATGIN\_DATASET$ function described in Algorithm \ref{alg:gen-data}. 
% We set the confidence parameter $\eta$ to $1e-3$, the restart number to $20,000$, and run PGD for 2,000 steps with a learning rate of $1e-2$.
% The dataset consists of 641 properties for the \Base model, 303 properties for the \Wide model, and 250 properties for the \Deep model.
% We also create a validation dataset with the same parameters used as for the test dataset on the \Base model consisting of 50 properties; we further create a training dataset with 2500 properties using $R=100$ restarts and running PGD for 1,000 steps.
        
% \import{Algorithms}{generating_dataset.tex}

\import{Sections/Appendix}{Network_Architecture}
\import{Sections/Appendix}{Generating_Dataset}
\import{Sections/Appendix}{GNN_Architecture}

\import{Sections/Appendix}{Running_Standard_Attacks}

\import{Sections/Appendix}{Baselines}
\clearpage
\newpage
\import{Sections/Appendix}{Further_Results}

\import{Sections/Appendix}{experiments_adv_trained}

%% file: Sections/Appendix/Network_Architecture.tex
\section{Network Architectures}\label{app:network_architecutres}

We now describe the three models used in this work in greater detail.
They have been trained robustly on the CIFAR-10 dataset \citep{krizhevsky2009learning} using the method introduced by \cite{wong2018provable} to achieve robustness against $l_\infty$ perturbations of size up to $\epsilon = 8/255$ (the amount typically considered in empirical works). The \Base and the \Wide model both have two convolutional layers, followed by two fully connected ones. The \Deep model has two further convolutional layers. All three networks use ReLU activations and all three models have been used in previous work \citep{Lu2020Neural, bunel2020lagrangian}.

\import{Tables}{network_architectures.tex}

%% file: Tables/network_architectures.tex
\begin{table}[h!]
	\centering
	\small
	\begin{tabular}{|c|c|c|}
		\hline
		\textbf{Network Name}\TBstrut & \textbf{No. of Properties} \TBstrut & \textbf{Network Architecture} \TBstrut\\
		\hline
		\begin{tabular}{l}
			\Base \\ 
			Model
		\end{tabular} & \begin{tabular}{@{}c@{}}
		    Training: 2500  \\
		    Validation: 50 \\
			Testing: 641
		\end{tabular} & \begin{tabular}{@{}c@{}} 
			\footnotesize
			Conv2d(3,8,4, stride=2, padding=1) \Tstrut \\
			Conv2d(8,16,4, stride=2, padding=1)\\
			linear layer of 100 hidden units \\
			linear layer of 10 hidden units\\
			%(Total ReLU activation units: 3172) \Bstrut
		\end{tabular}\\
		\hline
		\Wide Model & 303 & \begin{tabular}{@{}c@{}} 
			\footnotesize
			Conv2d(3,16,4, stride=2, padding=1) \Tstrut\\
			Conv2d(16,32,4, stride=2, padding=1)\\
			linear layer of 100 hidden units \\
			linear layer of 10 hidden units\\
			%(Total ReLU activation units: 6244) \Bstrut
		\end{tabular}\\
		\hline
		\Deep Model & 250 & \begin{tabular}{@{}c@{}} 
			\footnotesize
			Conv2d(3,8,4, stride=2, padding=1) \Tstrut \\
			Conv2d(8,8,3, stride=1, padding=1) \\
			Conv2d(8,8,3, stride=1, padding=1) \\
			Conv2d(8,8,4, stride=2, padding=1)\\
			linear layer of 100 hidden units \\
			linear layer of 10 hidden units\\
			%(Total ReLU activation units: 6756) \Bstrut
		\end{tabular}\\
		\hline
	\end{tabular}
	\caption{\label{tab:problem_size} Network Architectures.}%: Every layer apart from the last one is followed by a ReLU activation function.}%For each complete verification experiment, the network architecture used and the number of verification properties tested, from the dataset by \citet{lu2019neural}. Each layer but the last is followed by a ReLU activation function. For some of the property classes, we employed only a subset of the original data due to computational constraints.}
\end{table}

%% file: Sections/Appendix/Generating_Dataset.tex
\section{Generating the Dataset}\label{app:generating_dataset}

We generate a dataset for three different models: the \Base model, the \Wide model, and the \Deep model. For each of the three models we generate properties to attack, using the method described in Algorithm \ref{alg:gen-data}.
The algorithm runs binary search together with PGD-attack to find the smallest perturbation for each image for which there exists at least one adversarial example.
We generate a dataset setting the confidence parameter $\eta$ to $1e-3$, the restart number to $20,000$, and run PGD for 2,000 steps with a learning rate of $1e-2$.
We generate a dataset consisting of 641 properties for the \Base model, 303 properties for the \Wide model, and 250 properties for the \Deep model.
We also create a validation dataset with the same parameters used as for the test dataset on the \Base model consisting of 50 properties; we further create a training dataset also on the \Base model with 2500 properties using $R=100$ restarts and running PGD for 1,000 steps.\\\\\\\
        
\import{Algorithms}{generating_dataset.tex}

%% file: Algorithms/generating_dataset.tex
\begin{algorithm}[H]
\caption{Generating Dataset}\label{alg:gen-data}
\begin{algorithmic}[1]
\Function{Generating_Dataset}{$f, D, \eta, R, PGD\_hparams$}
% \Procedure{YourFunction}{$x$}
%   \State Do Something
% \EndProcedure
    \State{Provided: a trained network $f:\mathbb{R}^d \mapsto \mathbb{R}^m$, a set $D$ of $N$ pairs of images and their respective classes $(\x^i, y^i)$, a confidence parameter $\eta$, a restart parameter $R$, as well as parameters for PGD.}
    \For{$i = 1, \dots, N$}:
     \If{$\argmax f(\x^i) != y^i$}
         \State continue \Comment{If the network misclassifes the image, skip to the next one}
     \EndIf
     \State{$\ytar^i \gets \text{random number from }\{0, \cdots, m-1\}\setminus \{y^i\}$}
     \Comment{Pick a random incorrect class as target}
     \State{$\lb \gets  0$} \Comment{highest perturbation value for which we have failed to find an adversarial example}
     \State{$\ub \gets  0.5$} \Comment{lowest perturbation value for which we have found an adversarial example}
     \While{$\ub - \lb \geq \eta$}
     \State{$\epsilon^i \gets \frac{\lb + \ub }{2}$}
     \For{$j = 1, \dots, R$}
        \State{Run PGD with $(f, \x^i, y^i, \ytar^i, \epsilon^i)$} \Comment{Run PGD with $R$ restart or until found an adversarial example}
        \If{attack successful}
        \State{Break}
        \EndIf
     \EndFor
     \If{found adversarial example}
     \State{$\ub \gets \epsilon^i$} \Comment{Update $\ub$ as $\epsilon^i$ is now the lowest perturbation for which we have found an adversarial example}
     \Else
     \State{$\lb \gets \epsilon^i$} \Comment{Update $\lb$ as $\epsilon^i$ is now the highest perturbation for which we have failed to find an adversarial example}
     \EndIf
     \EndWhile
     \State{Record $(\x^i, y^i, \ytar^i, \epsilon^i)$}
    \EndFor
    \EndFunction
  \end{algorithmic}
\end{algorithm}

%% file: Sections/Appendix/GNN_Architecture.tex
\section{GNN Architecture}\label{app:GNN_architecture}
Having described the main structure of the GNN above, as well as the implementation of the forward and backward passes, and the final update step, we will now explain in greater detail how the node features are computed. The node features consist of three pieces of information: the gradient at the current point, the intermediate bounds of the neurons in the original network, and information from solving a standard  relaxation of the adversarial loss.
We now describe in greater detail how each of those parts is defined and computed.

\subsection{Intermediate Bounds}\label{app:sec:inter_bounds}
We recall the definition of the original network we are trying to attack:
$f(\x_0)= \hat{\x}_{L}\in {\mathbb{R}^m}$, where
\begin{align}
        \hat{\x}_{i+1} &= W^{i+1}\x_{i} + \bb^{i+1}, \qquad &&\text{for }i=0,\dots, L-1,\\
        \x_{i} &= \sigma(\hat{\x}_i), \qquad &&\text{for }i=1,\dots, L-1.
\end{align}
The adversarial problem can then be written as 
\begin{align}\label{eq:app:adv_prob}
        \min \hat{\x}_{L}[y] - \hat{\x}_{L}[\ytar] & && \\
        \hat{\x}_{i+1} &= W^{i+1}\x_{i} + \bb^{i+1}, \qquad &&\text{for }i=0,\dots, L-1,\\
        \x_{i} &= \sigma(\hat{\x}_i), \qquad &&\text{for }i=1,\dots, L-1,\\
        \x_0 &\in {\cal C} \subseteq {\mathbb{R}}^{d}
\end{align}
We now aim to compute bounds on the values that each neuron $\x_k[j]$ can take, where $k$ indexes the layer, and $j$ the neuron in that layer. The computation of the lower bound of a neuron can be described as finding a lower bound for the following minimization problem:
\begin{align}
        \min \hat{\x}_k[j] & && \\
        \hat{\x}_{i+1} &= W^{i+1}\x_{i} + \bb^{i+1}, \qquad &&\text{for }i=0,\dots, k-1,\\
        \x_{i} &= \sigma(\hat{\x}_i), \qquad &&\text{for }i=1,\dots, k-1,\\
        \x_0 &\in {\cal C} \subseteq {\mathbb{R}}^{d}.
\end{align}
We solve this using the method by \cite{wong2018provable} and using Interval Bound Propagation \citep{gowal2018effectiveness} and record the tighter of the two. We get the upper bound by changing the sign of the weights of the $k$-th layer function. We denote the lower and upper bounds for the $j$-th neuron in the $k$-th layer as $\lb_k[j]$ and $\ub_k[j]$, respectively.

\subsection{Solving a Standard Relaxation with Supergradient Ascent}\label{app:sec:dual}
We now describe a standard relaxation of the adversarial problem from the verification literature. Neural Network verification methods aim to solve the opposite problem of adversarial attacks. They try to prove that for a given network $f$, an image $\x$, a convex neighbourhood around it, $\mathcal{C}$, a true class $y$, and an incorrect target class $\ytar$, there does not exists an example $\x' \in \mathcal{C}$ that the network misclassifies as $\ytar$. In other words, it aims to show that no adversarial attack would be successful at finding an adversarial example.
This is equivalent to showing that the minimum in \eqref{eq:app:adv_prob} is strictly positive.

We now summarize the work of \cite{bunel2020lagrangian} who solve this problem using standard relaxations. First they relax the non-linear ReLU activation functions using the so-called Planet relaxation \citep{ehlers2017formal} before computing lower bounds using a formulation based on Lagrangian decompositions.

\paragraph{Planet Relaxation.} 
We denote the  output of the $k$-th layer before the application of the ReLU as $\zhat_k$ and the output of applying the ReLU to $\zhat_k$ as $\z_k$. Given the lower bounds $\lb_k$ and upper bounds $\ub_k$ of the values of $\zhat_k$, we relax the ReLU activations $\z_k = \sigma(\zhat_k)$ to its convex hull $\cvxhull$, defined as follows:
\begin{equation}
    \label{eq:cvx_hull}
    \cvxhull \equiv
    \begin{cases}
    \z_k[i] \geq 0 \;\;\; \z_k[i] \geq \zhat_k[i]\\
    \z_k[i] \leq \frac{\ubk[i](\zhat_k[i] - \lbk[i])}{\ubk[i] - \lbk[i]}  \quad &\iif \lbk[i] < 0 \aand \ubk[i] > 0\\
    \z_k[i] = 0													   &\iif \ubk[i] \leq 0 \\
    \z_k[i] = \zhat_k[i]										&\iif \lbk[i] \geq 0. \\
    \end{cases}
\end{equation}
%Note that the computation of the convex hull requires the knowing of the lower and upper bounds (i.e. $\lb_k$ and $\ub_k$) for each intermediate node. The bounds do not have to be optimal, however, the tighter the bounds are, the tighter the relaxation will be as well. 
% There are different ways of computing said bounds that have been proposed in the literature \cite{gowal2018effectiveness, raghunathan2018certified, wong2017provable}. In our experiments we use the method proposed by \citet{wong2017provable}.
To improve readability of our relaxation, we introduce the following notations for the constraints corresponding to the input and the $k$-th layer respectively:
\noindent\begin{minipage}{.5\linewidth}
\begin{equation*}
    \mathcal{P}_0(\z_0, \zhat_{1}) \equiv 
    \begin{cases}
    \z_0 \in C \\
    \zhat_1 = W_1 \z_0 + \bb_1 \\
    \end{cases}
\end{equation*}
\end{minipage}%
\begin{minipage}{.5\linewidth}
\begin{equation}
    \label{eq:P_constrains}
    \mathcal{P}_k(\zhat_{k}, \zhat_{k+1}) \equiv 
    \begin{cases}
    \exists \z_k \text{ s.t. }\\
    \lbk \leq \zhat_k \leq \ubk \\
    \cvxhull \\
    \zhat_{k+1} = W_{k+1} \z_k + \bb_{k+1}.
    \end{cases}
\end{equation}
\end{minipage}

Using the above notation, the Planet relaxation for computing the lower bound can be written as:
\begin{equation}\label{eq:primal}
    \min_{\z, \zhat} \zhat_n \text{ s.t. } \mathcal{P}_0(\z_0, \zhat_{1}); \mathcal{P}_k(\zhat_{k}, \zhat_{k+1}) \forkone.
\end{equation}

\paragraph{Lagrangian Decomposition.} We often merely need approximations of the bounds rather than the precise values of them: if we show that some valid lower bound of \eqref{eq:app:adv_prob} is strictly positive, then it follows that \eqref{eq:app:adv_prob} is also strictly positive and no adversarial example exists.
We can therefore make use of the primal-dual formulation of the problem as every feasible solution to the dual problem provides a valid lower bound for the primal problem.
Following the work of \citet{bunel2020lagrangian} we will use the Lagrangian decomposition \cite{guignard1987lagrangean}.
To this end, we first create two copies $\zhat_{A, k}, \zhat_{B, k}$ of each variable $\zhat_{k}$:
\begin{equation}\label{primal}
    \begin{aligned}
    \min_{\z, \zhat} \zhat_{A, n} \text{ s.t. } &\PO; \Pk &&\forkone\\
    &\zhat_{A, k} = \zhat_{B, k} &&\forkone.
    \end{aligned}
\end{equation}
Next we obtain the dual by introducing Lagrange multipliers $\rrho$ corresponding to the equality constraints of the two copies of each variable:
\begin{equation}\label{eq:dual}
    \begin{aligned}
    q(\rrho) = &\min_{\z, \zhat} &&\zhat_{A, n} + \sum_{k = 1, \dots, n-1} \rrho_k^\top (\zhat_{B, k} - \zhat_{A, k})\\
    &\text{ s.t. } &&\PO; \;\Pk \,\forkone.
    \end{aligned}
\end{equation}

\paragraph{Solving the Relaxation using Supergradient Ascent}
We solve the dual problem \eqref{eq:dual} using the supergradient ascent method proposed by \cite{bunel2020lagrangian}. We run supergradient ascent together with Adam for 100 steps to get a set of dual variables $\rrho$, as well as a matching set of primal variables $\x_0$ which, henceforth, we denote as $\x^{lp}$.

 \subsection{Node Features}
 For each node $\vv_k[i]$ we define a corresponding $q$-dimensional feature vector $\fki \in \mathbb{R}^q$
 describing the current state of that node. 
 \Pawandone{$d$ is overloaded.}
 We define the node features for the input layer as follows: 
 \begin{equation}
    \fkO := \left(\x^{t}[i], \sgn(\nabla_\x L(\x^t, y, y')[i]), \lb_0[i], \ub_0[i], \z^{lp}[i]\right)^{\top},
 \end{equation}
 and for the hidden and final layers as:
  \begin{equation}
    \fki := \left(\lb_k[i], \ub_k[i], \rrho_k[i] \right)^{\top}.
 \end{equation}
 Here, $\x^{t}$ is our current point, $\nabla_\x L(\x, y, y')$ is the gradient at the current point, and $\lb_k[i]$, and  $\ub_k[i]$ are the bounds for each node as described above (\S \ref{app:sec:inter_bounds}).
 Further, $\rrho_k$ is the current assignment to the corresponding dual variables computed using supergradient ascent and $\z^{lp}_k$ is the input corresponding to the primal solution of the dual (see \S \ref{app:sec:dual}).
 Other features can be used depending on the exact task or experimental setup. We note that there exists a trade-off between using more expressive features that are difficult to compute or simpler ones that are faster to compute.

 \subsection{Embeddings.}
For every node $v_k[i]$ we compute a corresponding $p$-dimensional embedding vector $\mmu_k[i] \in \RR^p$ using a learned function $g$:
\begin{equation}
    \mmu_k[i] := g (\fki).
\end{equation}
In our case $g$ is a simple multilayer perceptron (MLP), which is made up of a set of linear layers $\Theta_i$ and non-linear ReLU activations.
We train two different MLPs, one for the input layer, $g^{inp}$, and one for all other layers $g$.
We have the following set of trainable parameters:
% \begin{equation}
%     \Theta_0 \in \RR^{q \times p}, \quad
%     \Theta_1, \dots, \Theta_{T_1} \in \RR^{p \times p}, \quad
%     \bb_0, \dots, \bb_{T_1} \in \RR^{p}.
% \end{equation}
% \begin{equation}
%     \thetainp_0 \in \RR^{q \times p}, \quad
%     \thetainp_1, \dots, \thetainp_{T_1} \in \RR^{p \times p}, \quad
%     \bb_0, \dots, \bb_{T_1} \in \RR^{p}.
% \end{equation}
\begin{align}
\begin{split}
    &\thetainp_0 \in \RR^{5 \times p}, \quad
    \Theta_0 \in \RR^{3 \times p} \; \quad
    \thetainp_1, \dots, \thetainp_{T_1}, \;
    \Theta_1, \dots, \Theta_{T_1}  \in \RR^{p \times p}
\end{split}
\end{align}

Given feature vectors $\feat_0, \dots, \feat_L$ we compute the following set of vectors:
\begin{align}
    &\mmu^0_0 = \relu(\thetainp_0 \cdot \feat_0), \quad
    &&\mmu^{l+1}_0 = \relu(\thetainp_{l+1} \cdot \mmu^l_0), \quad &&&\text{ for } l = 1, \dots, T_1-1 &&&&\\
    &\mmu^0_k = \relu(\Theta_0 \cdot \feat_k), \quad
    &&\mmu^{l+1}_k = \relu(\Theta_{l+1} \cdot \mmu^l_k), \quad &&&\text{ for } l = 1, \dots, T_1-1; \; k = 1, \dots, L. &&&&
\end{align}
We initialize the embedding vector to be $\mmu_k = \mmu_k^{T_1}$, where $T_1+1$ is the depth of the MLP.

%% file: Sections/Appendix/Running_Standard_Attacks.tex
\section{Running Standard Algorithms Using A\lowercase{dv}GNN}\label{app:simulate_methods}

We show that our method is strictly more expressive than FGSM, I-FGSM, and PGD by showing that it can simulate each of them exactly.

FGSM aims to generate an adversarial example with the following update step:
\begin{equation}
    \x' = \x + \epsilon \sgn(\nabla_\x \advloss).
\end{equation}

 Let $\Theta_0$ be the zero-matrix with non-zero elements  $\Theta_0[{1,4}]=1$, $\Theta_0[{2,4}]=-1$.
 Moreover, setting $T_1=1$, $\Theta_1 = \eye$ and 
 $\bb_0 = \bb_1=\zero$, we get
 \begin{equation}\label{eq:feature_input}
    \fkO := \left(\x^{t}, \sgn(\nabla_\x L(\x^t, y, y')), \lb_k[i], \ub_k[i], \z^{lp}_k[i]\right)^{\top},
 \end{equation}
 
\begin{align}
     \mmu^0_k &= \left(\sgn(\nabla_\x L(\x^t, y, y')), -\sgn(\nabla_\x L(\x^t, y, y')), \zero, \dots, \zero \right)^{\top},\\
     \mmu &= \left( \left(\sgn(\nabla_\x L(\x^t, y, y'))\right)_+, -\left(\sgn(\nabla_\x L(\x^t, y, y'))\right)_-, \zero, \dots, \zero \right)^{\top}.
\end{align}

 If we set $\thetafor_2 = \thetafor_3 = \thetaback_2 = \thetaback_3 = \zero$ and $\thetafor_1 = \thetaback_1 = \eye$, then the forward and backward passes don't change the embedding vector.
 We now just need to set $\thetascore = (1, -1, 0, \dots, 0)^{\top}$ to get the new direction:
 
 \begin{equation}
    \dir = \thetascore \cdot \mmu_0 = \left(\sgn(\nabla_\x L(\x^t, y, y'))\right)_+ + \left(\sgn(\nabla_\x L(\x^t, y, y'))\right)_-
     = \sgn(\nabla_\x L(\x^t, y, y')).
\end{equation}

We now update as follows
\begin{equation}
    \x^{t+1} = \Pi_{\normball} \left( \x^t + \alpha \dir \right) = \Pi_{\normball} \left( \x^t + \alpha \sgn(\nabla_\x L(\x^t, y, y')) \right).
\end{equation}
Setting $\alpha = \epsilon$ we get the same update as FGSM.
We have shown that we can simulate FGSM using our GNN architecture by running AdvGNN once. Moreover, we can also simulate $T$ iterations of PGD or I-FGSM by running AdvGNN $T$ times.

%% file: Sections/Appendix/Baselines.tex
\section{Hyper-parameter Analysis for Baselines}
\subsection{PGD Attack}\label{app:hparam_pgd}
PGD aims to generate adversarial examples by picking $\x^0 \in \normball$ uniformly at random and then running the following update step for $T$ steps or until $L(\x^t, y, \ytar) > 0$:
\begin{equation}
    \x^{t+1} = \Pi_{\normball} \left( \x^t + \alpha \sgn(\nabla L(\x^t, y, \ytar) ) \right).
\end{equation}
We need to pick optimal values for the hyper-parameters $T$ and $\alpha$. We run a hyper-parameter analysis on the validation dataset described in section $\S \ref{app:generating_dataset}$. We try every combination of $T \in \{50, 100, 250, 1000\}$ and $\alpha \in \{1e-1, 1e-2, 1e-2\}$ and rank them both for the average time taken and the percentage of properties they time out. Taking the average of the two ranks we see that choosing $T=100$ and $\alpha = 0.01$ is the best combination (Table \ref{tab:hparam_pgd}).
% \import{Tables/Hparams}{hparam_pgd_val.tex}
We repeat the hyper-parameter on an easier version of the validation dataset which we get by adding a delta of 0.001 to the value of every perturbation. Just like for the original validation dataset, the following two combinations of hyper-parameters perform significantly better than all other combinations: ($T=1000, \alpha=0.001)$ and ($T=100, \alpha=0.01)$. They time out on the same number of properties but the former has a slightly lower average solving time this time.

\import{Tables/Hparams}{hparam_pgd_val.tex}

\newpage
\subsection{MI-FGSM+ Attack}\label{app:hparam_mi-fgsm}\label{app:MI-FGSM}
Adding momentum to the MI-FGSM attack was first suggested by \cite{dong2018boosting}. The original implementation is described in Algorithm \ref{alg:mi-fgsm}. This version does not perform well on our challenging dataset however. In fact it doesn't manage to find a single counter example on the validation dataset for any combination of hyper-parameters. 
One reason for this behaviour could be that often adversarial examples lie near the boundary of the input domain (at least in one dimension) and to reach those points every single update step needs to have the correct sign for that particular dimension (as we take $T$ steps of the form $\pm \sfrac{\epsilon}{T})$ . 
In order to improve its performance on difficult datasets we run it with random restarts. However, as the original implementation has no statistical elements, every run on the same image with the same hyper-parameters would have the same outcome. We thus adapt MI-FGSM to initialize the starting point uniformly at random from the input domain rather than starting at the original image. We further observed that initializing $\alpha$ as done in the original implementation greatly reduces its rate of success. We thus treat it as a hyper-parameter and give it as input to the function. We denote this optimized version of MI-FGSM as MI-FGSM+ and describe it in greater detail in Algorithm \ref{alg:mi-fgsm_opt}.
Similarly to PGD-Attack we now optimize over the hyper-parameters on the validation dataset. We try the following values: $T \in \{10, 100, 1000\}$, $\alpha \in \{1e-1, 1e-2, 1e-3\}$, $\eta \in \{0.0, 0.25, 0.5, 1.0\}$.
As we did for PGD we rank the performance of all combinations of hyper-parameters with respect to the number of properties successfully attack and average time taken (Table \ref{tab:hparam_mi_fgsm}). We get the following optimal set of hyper-parameters: $T=100, \alpha = 0.1, \eta=0.5$.

We also perform a similar analysis on an easier version of the validation dataset, where we add a constant (0.001) to the allowed perturbation value for each image. We reach the same optimal assignment for the three hyper-parameters as before.

\import{Algorithms}{MI_FGSM.tex}

\import{Tables/Hparams}{hparam_mifgsm.tex}
% \ref{alg:mi-fgsm}
% \import{Tables/Hparams}{hparam_mifgsm.tex}

\newpage
\subsection{Carlini and Wagner Attack}\label{app:hparam_CW}

We run the $l_\infty$ version of the Carlini and Wanger Attack ($C\%W$) \citep{carlini2017towards}. \CW aims to repeatedly optimize
\begin{equation}
    \min_{\delta} \; c \cdot h(x+ \delta) + \sum_i[(\delta_i - \tau)_+],
\end{equation}
for different values of $c$ and $\tau$, where h is a surrogate function based on the neural network we are trying to attack.
The method is described in greater detail in Algorithm \ref{alg:CW}. 

\CW has six hyper-parameters we search over: $T, c_{init}, c_{fin}, 
\gamma_{\tau}, \gamma_c, \alpha$.
Running every possible combination of assignments to the hyper-parameters like we did for PGD and MI-FGSM+ becomes computationally too expensive as the number of assignments increases exponentially in the number of parameters. Instead we split the search into three rounds. We initialize the parameters with those suggested in the original paper. In the first round we change one parameter at a time, keeping all other parameters constant. At the end of the first round we record the optimal values for each parameter. We evaluate the performance by taking the average of the minimum perturbation for which \CW managed to return a successful attack for each image.
We then repeat this process twice more: each time searching over the optimal hyper-parameter assignment one at a time, and updating the values at the end of each round.
At the end of the third round we reach the following assignment:
$T=100$, $c_{init}=1e-5$, $c_{fin}=1000$, 
$\gamma_{\tau}=0.99$, $\gamma_c=1.5$, $\alpha=1e-4$.\\

\import{Algorithms}{CW.tex}
\import{Tables/Hparams/}{hparam_CW.tex}

%% file: Tables/Hparams/hparam_pgd_val.tex
\begin{table}[h]
    \centering
    \caption{Hyper-parameter analysis for PGD attack on the Validation Set}\label{tab:hparam_pgd}
\begin{tabular}{rrrrrrr}
\toprule
$T$ &       $\alpha$  &  average\_time &  timeout &  rank\_time &  rank\_timeout &  average\_rank \\
\midrule
  100 &  0.01  &     87.740020 &            0.843137 &        1.0 &           2.0 &          1.50 \\
 1000 & 0.001  &     91.157906 &            0.862745 &        2.0 &           3.5 &          2.75 \\
  250 &  0.01  &     92.968972 &            0.823529 &        5.0 &           1.0 &          3.00 \\
  500 &  0.01  &     91.378347 &            0.862745 &        3.0 &           3.5 &          3.25 \\
 1000 &  0.01  &     91.607033 &            0.882353 &        4.0 &           5.0 &          4.50 \\
   50 &  0.01  &     93.659832 &            0.921569 &        6.0 &           6.5 &          6.25 \\
  500 & 0.001  &     94.735763 &            0.921569 &        7.0 &           6.5 &          6.75 \\
  100 &   0.1  &     99.852496 &            0.980392 &        8.0 &           8.0 &          8.00 \\
 1000 &   0.1  &    101.000000 &            1.000000 &       12.0 &          12.0 &         12.00 \\
  100 & 0.001  &    101.000000 &            1.000000 &       12.0 &          12.0 &         12.00 \\
  250 & 0.001  &    101.000000 &            1.000000 &       12.0 &          12.0 &         12.00 \\
  250 &   0.1  &    101.000000 &            1.000000 &       12.0 &          12.0 &         12.00 \\
  500 &   0.1  &    101.000000 &            1.000000 &       12.0 &          12.0 &         12.00 \\
   50 & 0.001  &    101.000000 &            1.000000 &       12.0 &          12.0 &         12.00 \\
   50 &   0.1  &    101.000000 &            1.000000 &       12.0 &          12.0 &         12.00 \\
\bottomrule
\end{tabular}
\end{table}

%% file: Algorithms/MI_FGSM.tex
\begin{algorithm}[H]
\caption{MI-FGSM \citep{dong2018boosting}}\label{alg:mi-fgsm}
\begin{algorithmic}[1]
\Function{MI-FGSM}{$f, \x, y, \ytar, \mu, T$}
    \State{$\alpha \gets \sfrac{\epsilon}{T}$} \Comment{Initialize stepsize parameter}
    \State{$\x^0 \gets \x$} \Comment{Initialize starting point}
    \State{$\g^0 \gets 0$} \Comment{Initialize momentum vector}
    \For{$t = 1, \dots, T$}:
        \State{$\g^{t+1} \gets \mu \cdot \g^t + \frac{\nabla_x L(\x^t, y, \ytar)}{\norm{\nabla_x L(\x^t, y, \ytar)}_1}$} \Comment{Update the momentum term}
        \State{$\x^{t+1} = \x^{t} + \alpha \cdot sgn(\g^{t+1})$} \Comment{Update the current point}
    \EndFor
    \State return $\x^T$
    \EndFunction
  \end{algorithmic}
\end{algorithm}

\begin{algorithm}[H]
\caption{MI-FGSM+}\label{alg:mi-fgsm_opt}
\begin{algorithmic}[1]
\Function{MI-FGSM+}{$f, \x, y, \ytar, \mu, T, \alpha$}
    \State{sample $\x^0$ from $\normball$} \Comment{Initialize starting point}
    \State{$\g^0 \gets 0$} \Comment{Initialize momentum vector}
    \For{$t = 1, \dots, T$}:
        \State{$\g^{t+1} \gets \mu \cdot \g^t + \frac{\nabla_x L(\x^t, y, \ytar)}{\norm{\nabla_x L(\x^t, y, \ytar)}_1}$} \Comment{Update the momentum term}
        \State{$\x^{t+1} = \Pi_{\normball} \left( \x^{t} + \alpha \cdot sgn(\g^{t+1}) \right)$} \Comment{Update the current point and project }
    \EndFor
    \State return $\x^T$
    \EndFunction
  \end{algorithmic}
\end{algorithm}

%% file: Tables/Hparams/hparam_mifgsm.tex
\begin{table}[h]
    \centering
    \caption{Hyper-parameter analysis for MI-FGSM+ on the Validation Set}\label{tab:hparam_mi_fgsm}
\begin{tabular}{rrrrrrrr}
\toprule
$T$ &    $\alpha$ &   $\mu$ &  average\_time &  timeout &  rank\_time &  rank\_timeout &  average\_rank \\
\midrule
  100 &   0.1 &  0.5 &     43.513870 &            0.305556 &        1.0 &           1.0 &          1.00 \\
  100 &   0.1 &  1.0 &     55.608030 &            0.500000 &        2.0 &           2.5 &          2.25 \\
 1000 &  0.01 &  0.5 &     59.396192 &            0.500000 &        3.0 &           2.5 &          2.75 \\
 1000 &   0.1 &  1.0 &     61.623909 &            0.527778 &        4.0 &           4.5 &          4.25 \\
 1000 &   0.1 &  0.5 &     63.214772 &            0.527778 &        5.0 &           4.5 &          4.75 \\
 1000 &  0.01 &  1.0 &     65.085730 &            0.583333 &        6.0 &           7.0 &          6.50 \\
  100 &   0.1 & 0.25 &     70.484347 &            0.555556 &        9.0 &           6.0 &          7.50 \\
 1000 &  0.01 & 0.25 &     67.918430 &            0.638889 &        7.0 &           8.5 &          7.75 \\
  100 &  0.01 &  0.5 &     69.199902 &            0.638889 &        8.0 &           8.5 &          8.25 \\
  100 &  0.01 & 0.25 &     75.356267 &            0.722222 &       10.0 &          10.5 &         10.25 \\
 1000 & 0.001 &  0.5 &     76.749888 &            0.722222 &       11.0 &          10.5 &         10.75 \\
 1000 & 0.001 & 0.25 &     82.939370 &            0.805556 &       12.0 &          12.5 &         12.25 \\
   10 &   0.1 &  0.5 &     83.524314 &            0.833333 &       13.0 &          14.5 &         13.75 \\
  100 &  0.01 &  1.0 &     83.739845 &            0.833333 &       14.0 &          14.5 &         14.25 \\
 1000 &   0.1 & 0.25 &     88.323196 &            0.805556 &       16.0 &          12.5 &         14.25 \\
 1000 & 0.001 &  1.0 &     87.845959 &            0.861111 &       15.0 &          16.0 &         15.50 \\
   10 &   0.1 &  1.0 &     90.158706 &            0.888889 &       17.0 &          17.0 &         17.00 \\
   10 &   0.1 & 0.25 &     94.782105 &            0.916667 &       18.0 &          18.0 &         18.00 \\
   10 &  0.01 & 0.25 &    100.012025 &            1.000000 &       19.0 &          23.0 &         21.00 \\
   10 & 0.001 &  0.5 &    100.012147 &            1.000000 &       20.0 &          23.0 &         21.50 \\
   10 & 0.001 &  1.0 &    100.012219 &            1.000000 &       21.0 &          23.0 &         22.00 \\
   10 &  0.01 &  1.0 &    100.012781 &            1.000000 &       22.0 &          23.0 &         22.50 \\
   10 &  0.01 &  0.5 &    100.013981 &            1.000000 &       23.0 &          23.0 &         23.00 \\
   10 & 0.001 & 0.25 &    100.015674 &            1.000000 &       24.0 &          23.0 &         23.50 \\
  100 & 0.001 &  1.0 &    100.119291 &            1.000000 &       25.0 &          23.0 &         24.00 \\
  100 & 0.001 &  0.5 &    100.124259 &            1.000000 &       26.0 &          23.0 &         24.50 \\
  100 & 0.001 & 0.25 &    100.134148 &            1.000000 &       27.0 &          23.0 &         25.00 \\
\bottomrule
\end{tabular}
\end{table}

%% file: Algorithms/CW.tex
\begin{algorithm}[H]
\caption{\CW}\label{alg:CW}
\begin{algorithmic}[1]
\Function{\CW}{$h, \x, y, \ytar, T, c_{init}, c_{fin}, 
\gamma_{\tau}, \gamma_c, \alpha$}
    \State{$c \gets c_{init}$}
    \State{$\tau \gets 1.0$}
    
    \While{$\tau < 0.1$ and $c < c_{fin}$}
        \State{
        \begin{equation}\label{eq:CW_in_alg}
            \min_{\delta} \; c \cdot h(x+ \delta) + \sum_i[(\delta_i - \tau)_+]
        \end{equation}
        }
        \State{Optimize \ref{eq:CW_in_alg} using the Adam optimizer with a learning rate of $\alpha$, and a step number of $T$}
        \If{found a counter example with $\delta_i \leq \tau \; \forall i$}
            \State{$\tau \gets \tau * \gamma_{\tau}$} \Comment{Decay $\tau$ using the decay factor $\gamma_{\tau}$}
            \State{$c \gets c * 1/2$} \Comment{Decay $c$ using factor $\gamma_c$}
        \Else
            \State{$c \gets c * \gamma_c$}
        \EndIf
    \EndWhile \State{}
    % \State{Return est $\delta$ found}
    \Return{Best $\delta$ found}
    \EndFunction
  \end{algorithmic}
\end{algorithm}

%% file: Tables/Hparams/hparam_CW.tex
\begin{table*}[h!]

%% local settings
\sisetup{detect-weight,mode=text}
% for avoiding siunitx using bold extended
\renewrobustcmd{\bfseries}{\fontseries{b}\selectfont}
\renewrobustcmd{\boldmath}{}
\newrobustcmd{\B}{\bfseries}

	\centering
	\scriptsize
	\setlength{\tabcolsep}{4pt}
	\aboverulesep = 0.1mm  % 0.605mm
	\belowrulesep = 0.2mm  % 0.984mm
    
	\begin{adjustbox}{center}

\begin{tabular}{lrrrrrrr}
\toprule
& $T$ &  $\alpha$ &  $c_{init}$ &  $c_{fin}$ &  $\gamma_{\tau}$ &  $\gamma_c$ &  Avg $(\epsilon_{val} - \epsilon_{\CW})$\\
\midrule
  Round 1 & 1000 &  1e-2 &    1e-5 &   20 &    0.9 &    2.0 &    - \\
\midrule
&   10 &  - &    - &   - &    - &    - &    -0.170515 \\
&  100 &  - &    - &   - &    - &    - &    \B -0.134574 \\
&   1000 &  - &    - &   - &    - &    - &   -0.146015 \\
&   - &  1e-3 &    - &   - &    - &    - &    \B  -0.050695  \\
&  - &  1e-2 &    - &   - &    - &    - &    -0.146015 \\
&  - &  1e-1 &    - &   - &    - &    - &    -0.717864 \\
&  - &  - &    1e-5 &   - &    - &    - &    -0.146015 \\
&  - &  - &    1e-4 &   - &    - &    - &    -0.140346 \\
&  - &  - &    1e-3 &   - &    - &    - &    \B -0.130972 \\
&  - &  - &    1e-2 &   - &    - &    - &    -0.149450 \\ 
&  - &  - &       - & 0.1 &    - &    - &    -0.199197 \\
&  - &  - &       - &   1 &    - &    - &    -0.197057 \\
&  - &  - &       - &  10 &    - &    - &    -0.160842 \\
&  - &  - &       - & 100 &    - &    - &    \B -0.105023 \\
&  - &  - &       - &     &  0.5 &    - &    -0.221801 \\
&  - &  - &       - &     &  0.9 &    - &    \B -0.146015 \\
&  - &  - &       - &     & 0.99 &    - &    -0.180077 \\  
&  - &  - &       - &     &    - &  1.5 &    -0.161380 \\
&  - &  - &       - &     &    - &  2.0 &    \B -0.146015 \\
&  - &  - &       - &     &    - &  5.0 &    -0.162026 \\
\midrule
  Round 2 & 100 &  1e-3 &    1e-3 &  100 &    0.9 &    2.0 &    - \\
\midrule
&   10 &  - &    - &   - &    - &    - &    -0.068567 \\
&  100 &  - &    - &   - &    - &    - &    \B  -0.051525 \\
&   1000 &  - &    - &   - &    - &    - &   -0.052210 \\
&   - &  1e-4 &    - &   - &    - &    - &    -0.059814  \\
&  - &  1e-3 &    - &   - &    - &    - &    \B   -0.051525 \\
&  - &  1e-2 &    - &   - &    - &    - &    -0.101191 \\
&  - &  - &    1e-5 &   - &    - &    - &    -0.051074 \\
&  - &  - &    1e-4 &   - &    - &    - &    \B  -0.050897 \\
&  - &  - &    1e-3 &   - &    - &    - &    -0.051525\\
&  - &  - &    1e-2 &   - &    - &    - &     -0.053127 \\ 
&  - &  - &       - &  10 &    - &    - &    -0.053124 \\
&  - &  - &       - & 100 &    - &    - &    -0.051525 \\
&  - &  - &       - & 1000 &    - &    - &    \B -0.051488 \\
&  - &  - &       - &     &  0.5 &    - &    -0.180116 \\
&  - &  - &       - &     &  0.9 &    - &    -0.051525 \\
&  - &  - &       - &     & 0.99 &    - &    \B  -0.036093 \\  
&  - &  - &       - &     &    - &  1.5 &    \B  -0.051445 \\
&  - &  - &       - &     &    - &  2.0 &    -0.051525\\
&  - &  - &       - &     &    - &  5.0 &    -0.052622 \\
\midrule
  Round 3 & 100 &  1e-3 &    1e-4 &  1000 &    0.99 &    1.5 &    - \\
\midrule
&   10 &  - &    - &   - &    - &    - &    -0.045861 \\
&  100 &  - &    - &   - &    - &    - &    \B  -0.035893 \\
&   1000 &  - &    - &   - &    - &    - &   -0.101963 \\
&   - &  1e-4 &    - &   - &    - &    - &   \B   -0.033943  \\
&  - &  1e-3 &    - &   - &    - &    - &     -0.035893 \\
&  - &  1e-2 &    - &   - &    - &    - &    -0.098903 \\
&  - &  - &    1e-5 &   - &    - &    - &    \B  -0.035488 \\
&  - &  - &    1e-4 &   - &    - &    - &    -0.035893 \\
&  - &  - &    1e-3 &   - &    - &    - &    0.035676\\
&  - &  - &       - &  10 &    - &    - &    -0.035957 \\
&  - &  - &       - & 100 &    - &    - &    \B -0.035893 \\
&  - &  - &       - & 1000 &    - &    - &    \B -0.035893 \\
&  - &  - &       - &     &  0.9 &    - &    -0.049217 \\
&  - &  - &       - &     &  0.99 &    - &    \B -0.035893 \\
&  - &  - &       - &     & 0.999 &    - &     -0.128462 \\  
&  - &  - &       - &     &    - &  1.25 &    -0.037318 \\
&  - &  - &       - &     &    - &  1.5 &    \B  -0.035893\\
&  - &  - &       - &     &    - &  2.0 &    -0.037543\\
\bottomrule
\end{tabular}
	\end{adjustbox}
    \caption{Hyper-parameter analysis for \CW attack on the Validation Set}\label{tab:hparam_CW}
\end{table*}

%% file: Sections/Appendix/Further_Results.tex
% \section{Experimental Setup}
% \subsection{Training}

% \subsection{Fine-Tuning}
% \subsection{Experiments}

\section{Further Experimental Results}\label{app:further_results}
\subsection{Main Experiments}

All methods apart from \CW use random initialization. We therefore run every experiment in this paper three times, each time with a different random seed (using the Pytorch implementation of random seeds).
We manually set the time taken to 100 if a method times out on a property.
We summarize the results in Table \ref{table:main_seeds} and Figure \ref{fig:cactus_main_seeds}. We can see that even though the random seed makes a significant different for a single attack, when taking the average over the entire dataset the differences are very small. In particular, the difference between the results for the same attack with different seeds is much smaller than the difference between methods. This shows that our results are statistically significant.

\import{Tables}{main_exp_seeds.tex}
\import{Figures}{cactus_plots_seeds.tex}
\subsection{Easy Experiments}
% Mention that traind on different GPUS

As mentioned above, we also run experiments on an easier dataset. In practice there may be use cases where we want to generate easy adversarial examples very quickly, hence it is beneficial for strong methods to also work well on easier tasks.
As all methods will generate adversarial examples more quickly on this easier version of the dataset we reduce the timeout to 20 seconds.
The results are summarized in Tables \ref{table:easy_seeds_20} and \ref{table:easy_combined_20} and Figure \ref{fig:cactus_easy_seeds}. AdvGNN outperforms all baselines on all three models. On the \Base model in particular we reduce the percentage of properties on which our method times out by over 98\% compared to each of the three baselines. When comparing the results for the different seeds we see that every single run of AdvGNN beat every other run of any of the baselines, again showing that changing the random seed does not change the outcome significantly.

% \import{Tables}{easy_exp_seeds.tex}
\import{Tables}{easy_exp_seeds_timeout20.tex}
% \import{Tables}{easy_exp_combined}
\import{Tables}{easy_exp_combined_timeout20}
\import{Figures}{cactus_plots_easy_seeds.tex}

%% file: Tables/main_exp_seeds.tex
% \subsection{Tables in process}
%  \sisetup{detect-weight=true,detect-inline-weight=math,detect-mode=true}
\begin{table*}[h]

%% local settings
\sisetup{detect-weight,mode=text}
% for avoiding siunitx using bold extended
\renewrobustcmd{\bfseries}{\fontseries{b}\selectfont}
\renewrobustcmd{\boldmath}{}
% abbreviation
\newrobustcmd{\B}{\bfseries}

	\centering
	\scriptsize
	\setlength{\tabcolsep}{4pt}
	\aboverulesep = 0.1mm  % 0.605mm
	\belowrulesep = 0.2mm  % 0.984mm
	
	% bold elements in table
	%\renewrobustcmd{\bfseries}{\fontseries{b}\selectfont}
    %\renewrobustcmd{\boldmath}{}
    %\newrobustcmd{\B}{\bfseries}
    
	\begin{adjustbox}{center}

\begin{tabular}{llrrrrrr}
			& & \multicolumn{2}{ c }{\Base Model} & \multicolumn{2}{ c }{\Wide Model} & \multicolumn{2}{ c }{\Deep Model} \\
			\toprule

% \toprule
     Method &    Seed    & Time(s) &  Timeout(\%) &  Time(s) &  Timeout(\%) &  Time(s) &  Timeout(\%) \\

    \cmidrule(lr){1-1} \cmidrule(lr){2-2} \cmidrule(lr){3-4} \cmidrule(lr){5-6} \cmidrule(lr){7-8}
    
% \midrule
 PGD Attack &  2222 &   87.354 &      82.995 &        80.542 &           74.917 &        83.764 &             79.2 \\
 PGD Attack &  3333 &   87.396 &      83.151 &        80.301 &           75.908 &        84.930 &             81.2 \\
 PGD Attack &  4444 &   87.488 &      82.839 &        80.404 &           75.248 &        84.355 &             81.2 \\
     MI-FGSM+ &  2222 &   39.897 &      26.677 &        31.583 &           21.122 &        59.887 &             46.4 \\
    MI-FGSM+ &  3333 &   39.763 &      26.053 &        30.761 &           20.462 &        61.380 &             49.2 \\
    MI-FGSM+ &  4444 &   41.655 &      28.705 &        31.087 &           19.802 &        60.467 &             48.0 \\
 C\&W &     0 &   97.385 &      95.164 &        96.366 &           93.729 &        99.321 &             97.6 \\
     AdvGNN &  2222 &   14.152 &      10.296 &        24.429 &           18.812 &        52.337 &             43.6 \\
     AdvGNN &  3333 &   \B12.937 &       \B8.580 &        \B23.501 &           \B17.822 &        \B50.054 &             \B42.0 \\
     AdvGNN &  4444 &   13.490 &       9.360 &        24.338 &           18.812 &        52.616 &             44.0 \\

\bottomrule
\end{tabular}

	\end{adjustbox}
	\caption{\small We compare average (mean) solving time and the percentage of properties that the methods time out on when using a cut-off time of 100s and the random Pytorch seeds specified. The best performing method for each subcategory is highlighted in bold. AdvGNN is the best performing method as every single run of AdvGNN beats every other run by any of the other methods on each model.}
	\label{table:main_seeds}
\end{table*}

%% file: Figures/cactus_plots_seeds.tex
 \begin{figure*}[h]
	\centering
	\begin{subfigure}{.32\textwidth}
		\centering
		\includegraphics[width=\textwidth]{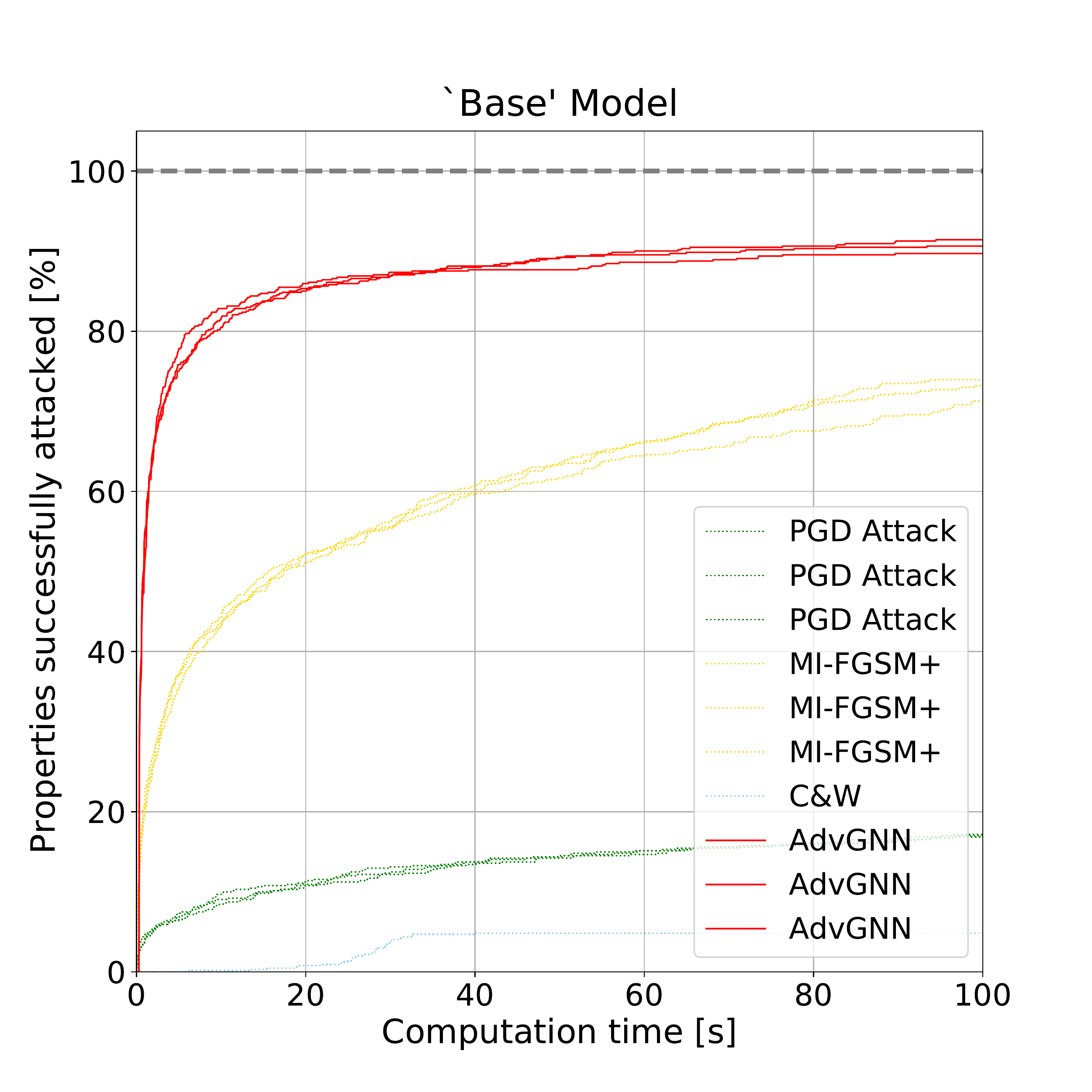}
		%\subcaption{subfigure}{680 steps of subgradient vs. 1040 steps of Dvijotham}
		%\label{fig:600prox}
	\end{subfigure}
	\begin{subfigure}{.32\textwidth}
		\centering
		\includegraphics[width=\textwidth]{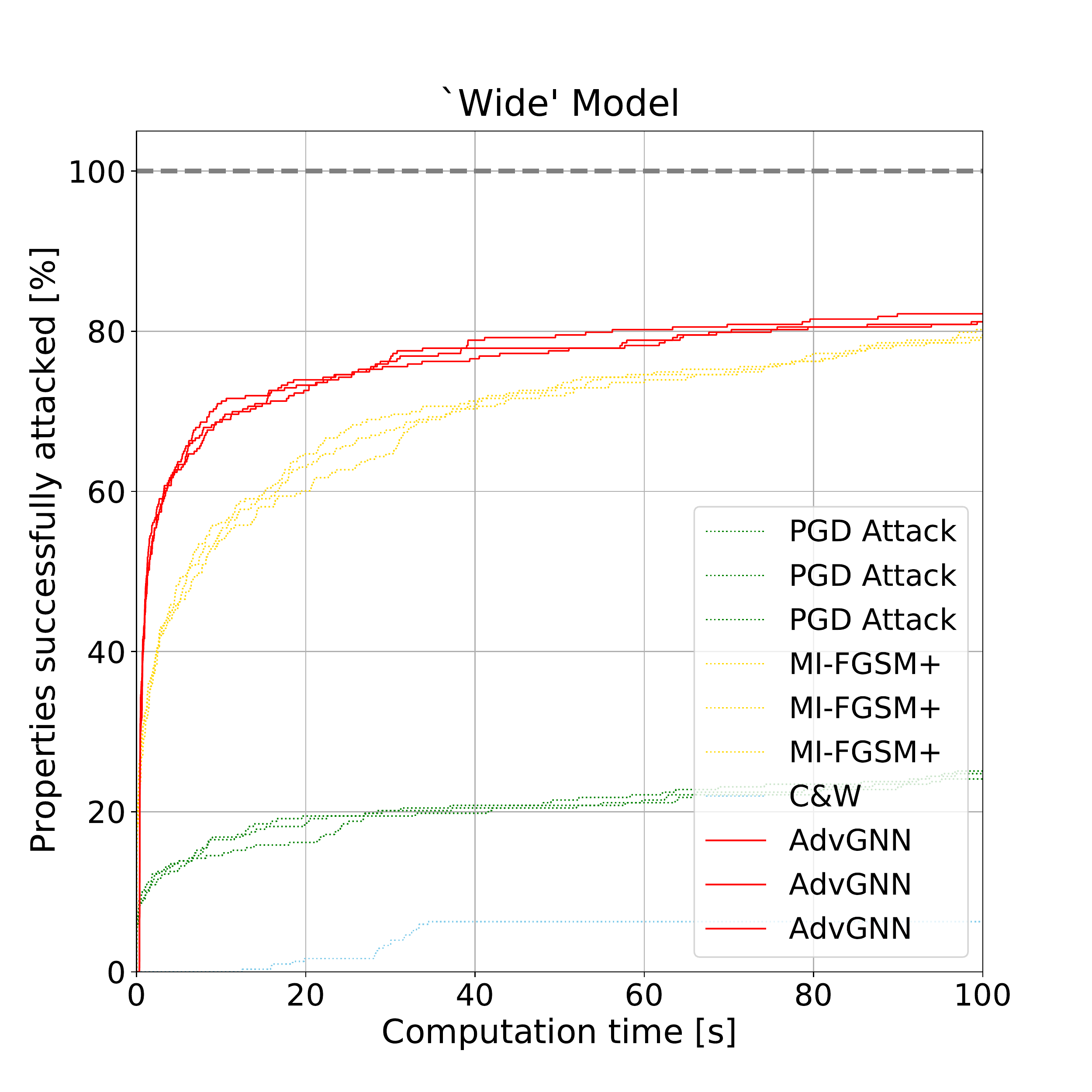}
		%\subcaption{400 steps of proximal vs. 680 steps of subgradient}
		%$label{fig:300prox} 
	\end{subfigure}
	\begin{subfigure}{.32\textwidth}
		\centering
		\includegraphics[width=\textwidth]{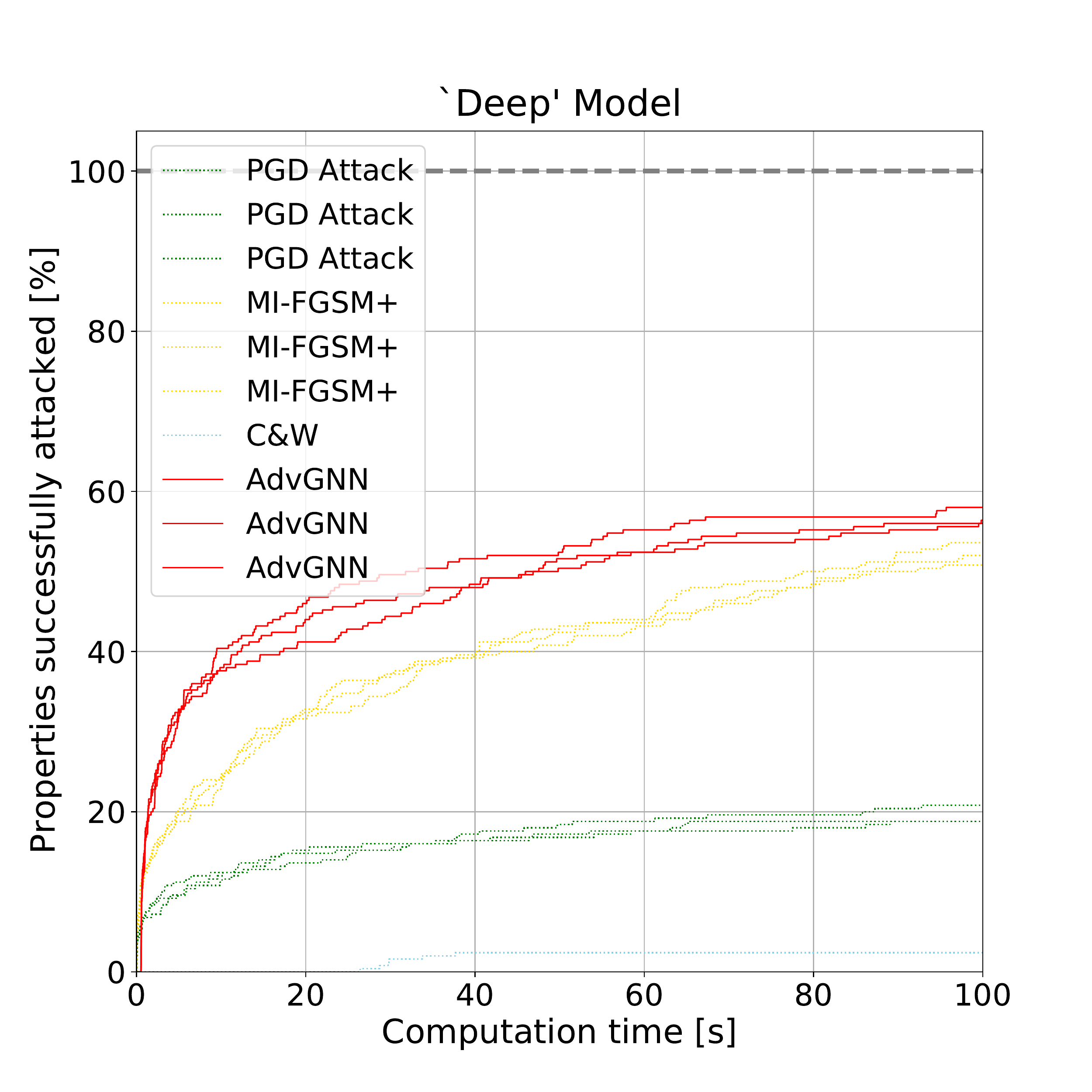}
		%\subcaption{400 steps of proximal vs. 680 steps of subgradient}
		%$label{fig:300prox} 
	\end{subfigure}
	\caption{Cactus plots for the main datasets on the \Base, \Wide and \Deep models. For each, we compare the attack methods by plotting the percentage of successfully attacked images as a function of runtime.}
	\label{fig:cactus_main_seeds}
\end{figure*}

%% file: Tables/easy_exp_seeds_timeout20.tex
% \subsection{Tables in process}
%  \sisetup{detect-weight=true,detect-inline-weight=math,detect-mode=true}
\begin{table*}[h]

%% local settings
\sisetup{detect-weight,mode=text}
% for avoiding siunitx using bold extended
\renewrobustcmd{\bfseries}{\fontseries{b}\selectfont}
\renewrobustcmd{\boldmath}{}
% abbreviation
\newrobustcmd{\B}{\bfseries}

	\centering
	\scriptsize
	\setlength{\tabcolsep}{4pt}
	\aboverulesep = 0.1mm  % 0.605mm
	\belowrulesep = 0.2mm  % 0.984mm
	
	% bold elements in table
	%\renewrobustcmd{\bfseries}{\fontseries{b}\selectfont}
    %\renewrobustcmd{\boldmath}{}
    %\newrobustcmd{\B}{\bfseries}
    
	\begin{adjustbox}{center}

\begin{tabular}{llrrrrrr}
			& & \multicolumn{2}{ c }{\Base-Easy} & \multicolumn{2}{ c }{\Wide-Easy} & \multicolumn{2}{ c }{\Deep-Easy} \\
			\toprule

% \toprule
     Method &    Seed    & Time(s) &  Timeout(\%) &  Time(s) &  Timeout(\%) &  Time(s &  Timeout(\%) \\

    \cmidrule(lr){1-1} \cmidrule(lr){2-2} \cmidrule(lr){3-4} \cmidrule(lr){5-6} \cmidrule(lr){7-8}
    
% \midrule

    PGD Attack &  2222 &    4.698 &      15.445 &         2.509 &            7.261 &         4.166 &             11.2 \\
    PGD Attack &  3333 &    4.714 &      14.353 &         2.109 &            5.611 &         3.655 &              8.0 \\
    PGD Attack &  4444 &    4.719 &      15.133 &         2.830 &            9.571 &         4.073 &             11.2 \\
    MI-FGSM+ &  2222 &    1.123 &       2.340 &         0.810 &            1.320 &         1.703 &              4.0 \\
    MI-FGSM+ &  3333 &    1.398 &       3.432 &         0.712 &            0.660 &         1.570 &              2.0 \\
    MI-FGSM+ &  4444 &    1.343 &       3.120 &         0.813 &            0.990 &         1.461 &              2.8 \\
    C\&W &     0 &   17.030 &      69.111 &        15.978 &           60.396 &        17.487 &             76.0 \\
    AdvGNN &  2222 &    0.509 &       0.156 &         \B0.550 &            \B0.330 &         1.443 &              \B0.8 \\
    AdvGNN &  3333 &    \B0.505 &       \B0.000 &         0.569 &            \B0.330 &         \B1.351 &              \B0.8 \\
    AdvGNN &  4444 &    0.538 &       \B0.000 &         0.665 &            \B0.330 &         1.603 &              1.2 \\

\bottomrule
\end{tabular}

	\end{adjustbox}
	\caption{\small We compare average (mean) solving time and the percentage of properties that the methods time out on when using a cut-off time of 20s and the random Pytorch seeds specified. The best performing method for each subcategory is highlighted in bold.}
	\label{table:easy_seeds_20}
\end{table*}

%% file: Tables/easy_exp_combined_timeout20.tex
% \subsection{Tables in process}
%  \sisetup{detect-weight=true,detect-inline-weight=math,detect-mode=true}
\begin{table*}[h]

%% local settings
\sisetup{detect-weight,mode=text}
% for avoiding siunitx using bold extended
\renewrobustcmd{\bfseries}{\fontseries{b}\selectfont}
\renewrobustcmd{\boldmath}{}
% abbreviation
\newrobustcmd{\B}{\bfseries}

	\centering
	\scriptsize
	\setlength{\tabcolsep}{4pt}
	\aboverulesep = 0.1mm  % 0.605mm
	\belowrulesep = 0.2mm  % 0.984mm
	
	% bold elements in table
	%\renewrobustcmd{\bfseries}{\fontseries{b}\selectfont}
    %\renewrobustcmd{\boldmath}{}
    %\newrobustcmd{\B}{\bfseries}
    
	\begin{adjustbox}{center}

\begin{tabular}{lrrrrrr}
			& \multicolumn{2}{ c }{\Base-Easy} & \multicolumn{2}{ c }{\Wide-Easy} & \multicolumn{2}{ c }{\Deep-Easy} \\
			\toprule

% \toprule
     Method &    Time(s) &  Timeout(\%) &  Time(s) &  Timeout(\%) &  Time(s) &  Timeout(\%) \\

    \cmidrule(lr){1-1} \cmidrule(lr){2-3} \cmidrule(lr){4-5} \cmidrule(lr){6-7}

 PGD Attack &    4.710 &      14.977 &         2.483 &            7.481 &         3.965 &           10.133 \\
 MI-FGSM+ &    1.288 &       2.964 &         0.778 &            0.990 &         1.578 &            2.933 \\
 C\&W &   17.030 &      69.111 &        15.978 &           60.396 &        17.487 &           76.000 \\
 AdvGNN &    \B0.518 &       \B0.052 &         \B0.595 &            \B0.330 &         \B1.465 &            \B0.933 \\

\bottomrule
\end{tabular}

	\end{adjustbox}
	\caption{\small We compare average (mean) solving time and the percentage of properties that the methods time out on when using a cut-off time of 20s. The best performing method for each subcategory is highlighted in bold.}
	\label{table:easy_combined_20}
\end{table*}

%% file: Figures/cactus_plots_easy_seeds.tex
 \begin{figure*}[h!]
	\centering
	\begin{subfigure}{.32\textwidth}
		\centering
		\includegraphics[width=\textwidth]{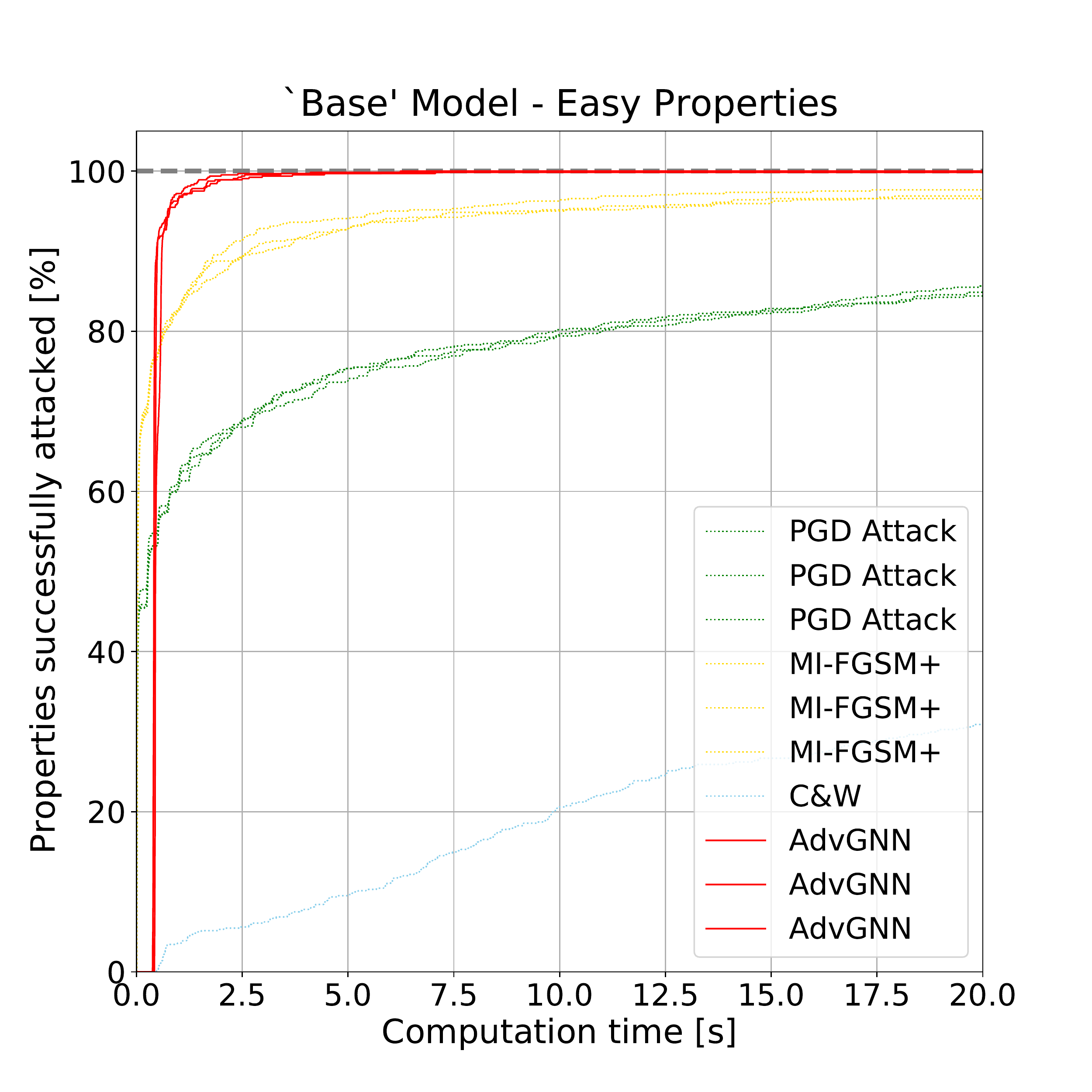}
		%\subcaption{subfigure}{680 steps of subgradient vs. 1040 steps of Dvijotham}
		%\label{fig:600prox}
	\end{subfigure}
	\begin{subfigure}{.32\textwidth}
		\centering
		\includegraphics[width=\textwidth]{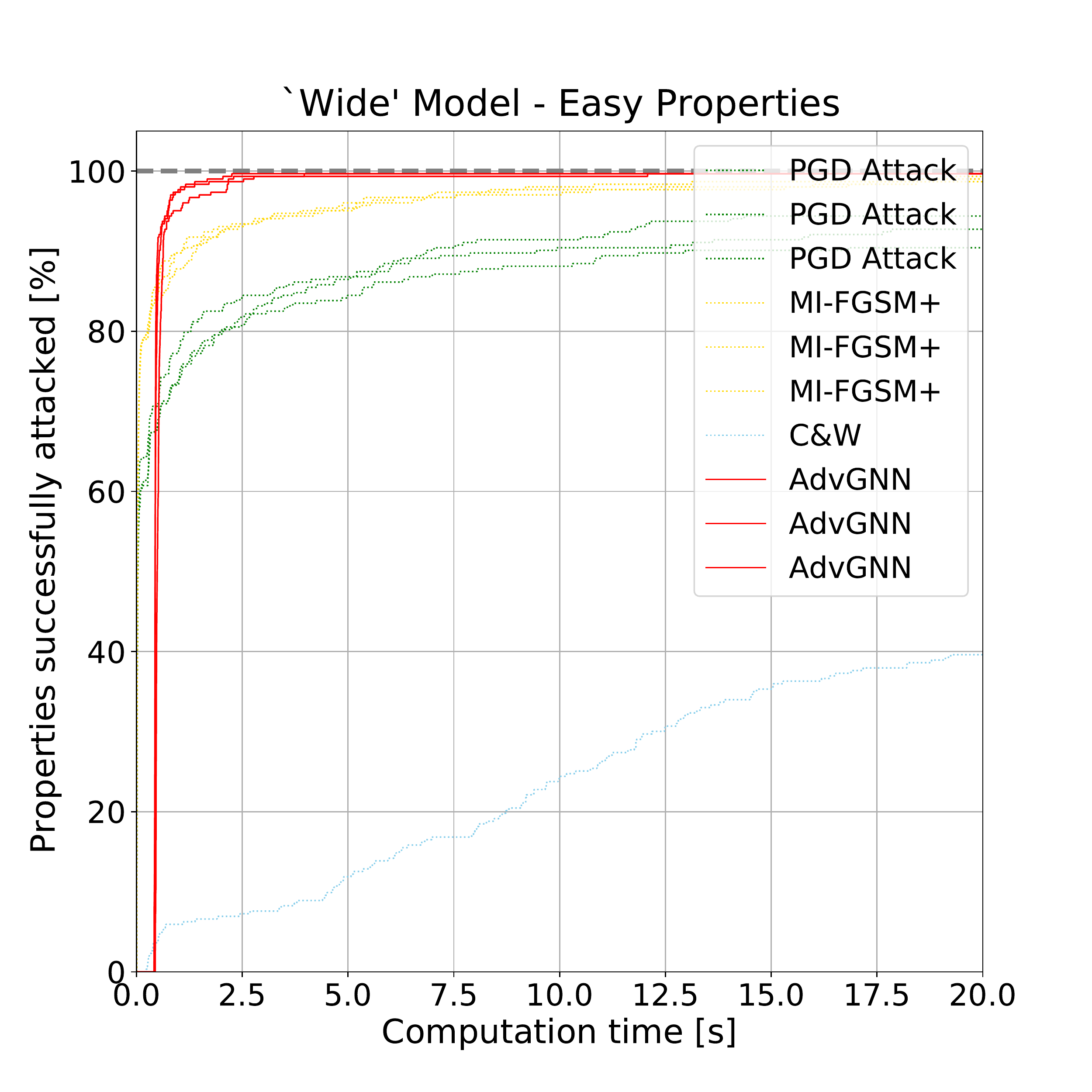}
		%\subcaption{400 steps of proximal vs. 680 steps of subgradient}
		%$label{fig:300prox} 
	\end{subfigure}
	\begin{subfigure}{.32\textwidth}
		\centering
		\includegraphics[width=\textwidth]{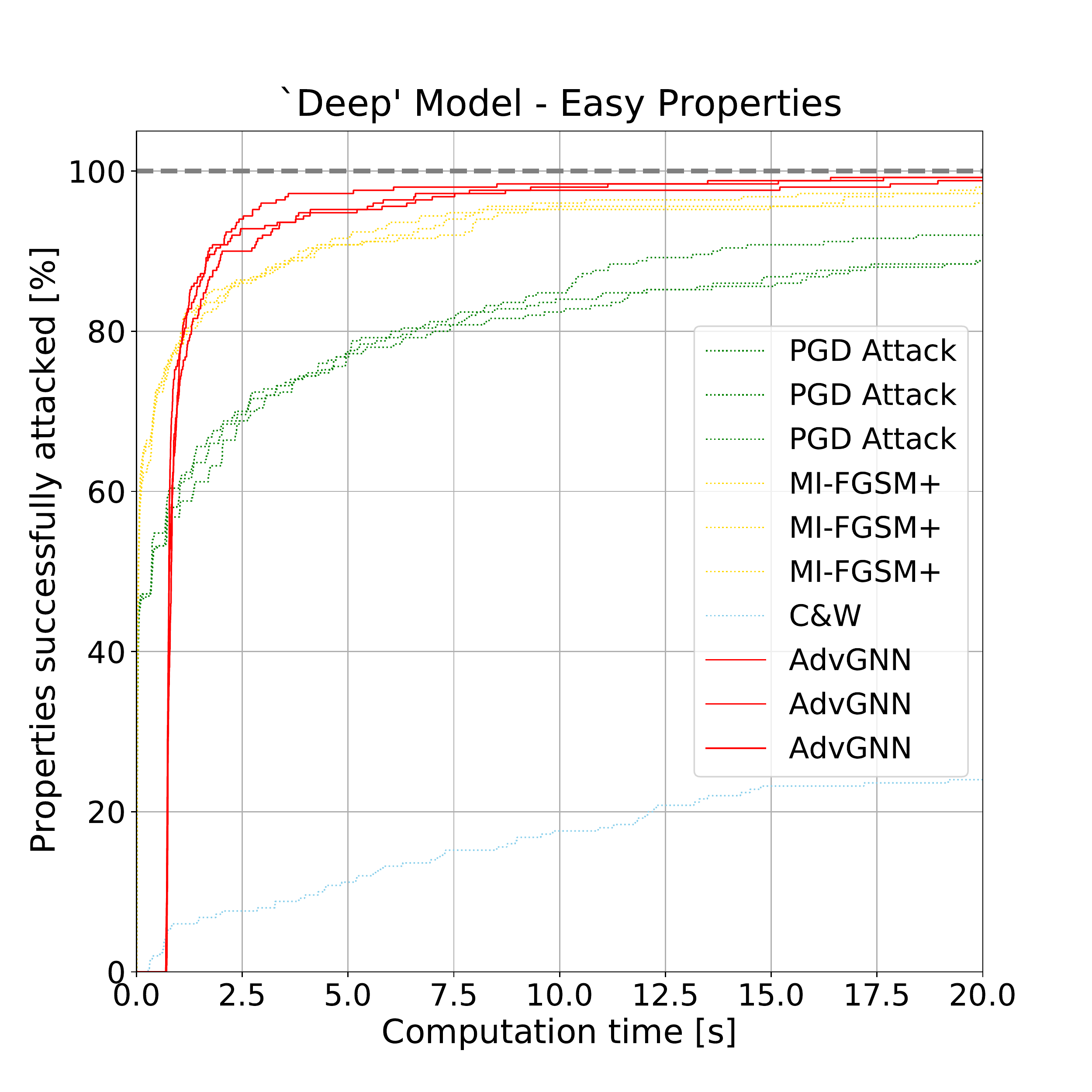}
		%\subcaption{400 steps of proximal vs. 680 steps of subgradient}
		%$label{fig:300prox} 
	\end{subfigure}
	\caption{Cactus plots for the easy datasets on the \Base, \Wide and \Deep models. For each, we compare the attack methods by plotting the percentage of successfully attacked images as a function of runtime.}
	\label{fig:cactus_easy_seeds}
\end{figure*}

%% file: Sections/Appendix/experiments_adv_trained.tex
\subsection{Experiments on an adversarially trained model}

We can confirm that our approach also works for adversarially trained models. We train a neural network that has the same artchitecture as the \Wide model used above using the method by \cite{madry2018towards}. After finetuning our GNN on an this adversarially trained CIFAR10 model, advGNN outperforms both PGD and MI-FGSM+. We run all three methods on 101 properties with a timeout of 20 seconds and repeat the experiment three times with three different random seeds. AdvGNN clearly outperforms both baselines timing out on 14\% of all properties compared to 21\% for MI-FGSM+ and 78\% for PGD, reducing average solving time by over 30\% (see Table \ref{table:adv_model}.
% \import{Tables/}{adv_model.tex}
\import{Tables/}{adv_model2.tex}

\subsection{Ablation study -  simpled feature vectors}
Computing the features vectors (Equations (32) and (33)) requires solving a linear program (Equation (31)). However, if we use a simpler approach as proposed by Kolter and Wong (2018) instead of super-gradient ascent our method still outperforms all baselines, successfully attacking 86\% of all properties on the base model compared to 5\%, 17\%, and 73\% for the three baselines, respectively (Table \ref{tab:kw_naive_features}). The reduced performance compared to the original AdvGNN performance shows that the feature vector plays a significant role in generating better directions. At the same time the modified AdvGNN method still outperforms all baselines indicating that the KW can be used when we run our method on larger networks.
\import{Tables/}{kw_naive_features.tex}

%% file: Tables/adv_model2.tex
\begin{table}[h]
    \centering

    \begin{tabular}{rlll}
      \toprule % from booktabs package
      \bfseries Method & \bfseries Seed & \bfseries Time(s) & \bfseries Timeout(\%) \\
      \midrule % from booktabs package
 PGD Attack &  2222 &   16.922 &  79.2  \\
 PGD Attack &  3333 &   16.222 &  78.2  \\
 PGD Attack &  4444 &   16.382 &  77.2 \\
    MI-FGSM+ &  2222 &    5.771 &  27.8    \\
    MI-FGSM+ &  3333 &     5.773  &  18.8 \\
    MI-FGSM+ &  4444 &    5.847 &  20.8  \\
     AdvGNN &  2222 &    4.079 &  12.9   \\
     AdvGNN &  3333 &    3.739  & 12.9  \\
     AdvGNN &  4444 &    3.851 &  14.9 \\
      \bottomrule % from booktabs packag
    \end{tabular}
	\caption{\small We run experiments on the adversarially trained \Wide model. We compare average (mean) solving time and the percentage of properties that the methods time out on when using a cut-off time of 20s and the random Pytorch seeds specified.}
	\label{table:adv_model}
\end{table}

%% file: Tables/kw_naive_features.tex
\begin{table}[h]
    \centering

    \begin{tabular}{rll}
      \toprule % from booktabs package
      \bfseries Method & \bfseries Time(s) & \bfseries Timeout(\%) \\
      \midrule % from booktabs package
 PGD Attack &   87.412 &      82.995        \\ 
 MI-FGSM+ &   40.438 &      27.145     \\
 C\&W &   97.385 &      95.164 \\
 AdvGNN-s & 19.788 & 13.885 \\
 AdvGNN &   \bfseries 13.527 &       \bfseries 9.412   \\
      \bottomrule % from booktabs packag
    \end{tabular}
    \caption{\Base Model. We compare average (mean) solving time and the percentage of properties that the methods time out on when using a cut-off time of 100s. AdvGNN is the main method described; AdvGNN-s uses the simple KW method rather than the iterative supergradient ascent method to compute the feature vector (\ref{eq:feature_input})}\label{tab:kw_naive_features}
\end{table}